\documentclass[runningheads]{llncs}

\usepackage{eccv}

\usepackage{eccvabbrv}

\usepackage{graphicx}
\usepackage{booktabs}

\usepackage[accsupp]{axessibility}  

\usepackage{hyperref}

\usepackage{orcidlink}

\usepackage{bbm}
\usepackage{booktabs}
\usepackage{multirow}
\usepackage{caption}
\usepackage{float}
\begin{document}

\title{Towards Adaptive Pseudo-label Learning for Semi-Supervised Temporal Action Localization} 

\titlerunning{Adaptive Pseudo-label Learning 
 for SS-TAL}

\author{Feixiang  Zhou\orcidlink{0000-0003-4939-9393} \and
Bryan Williams\orcidlink{0000-0001-5930-287X} \and
Hossein Rahmani\thanks{Corresponding author}\orcidlink{0000-0003-1920-0371}} 

\authorrunning{F.~Zhou et al.}

\institute{Lancaster University, UK  
\\
\email{\{f.zhou3,b.williams6, h.rahmani\}@lancaster.ac.uk}}

\maketitle

\begin{abstract}
Alleviating noisy pseudo labels remains a key challenge in Semi-Supervised Temporal Action Localization (SS-TAL).  Existing methods often filter pseudo labels based on strict conditions, but they typically assess classification and localization quality separately, leading to suboptimal pseudo-label ranking and selection. In particular, there might be inaccurate pseudo labels within selected positives, alongside reliable counterparts erroneously assigned to negatives. To tackle these problems, we propose a novel Adaptive Pseudo-label Learning (APL) framework to facilitate better pseudo-label selection. Specifically, to improve the ranking quality, Adaptive Label Quality Assessment (ALQA) is proposed to jointly learn classification confidence and localization reliability, followed by dynamically selecting pseudo labels based on the joint score. Additionally, we propose an Instance-level Consistency Discriminator (ICD) for eliminating ambiguous positives and mining potential positives simultaneously based on inter-instance intrinsic consistency, thereby leading to a more precise selection. We further introduce a general unsupervised Action-aware Contrastive Pre-training (ACP) to enhance the discrimination both within actions and between actions and
backgrounds, which benefits SS-TAL. Extensive experiments on THUMOS14 and ActivityNet v1.3 demonstrate that our method achieves state-of-the-art performance under various semi-supervised settings. 
\end{abstract}

\section{Introduction}
\label{sec:intro}

Temporal action localization (TAL) aims to localize temporal boundaries of action instances and identify corresponding categories from an untrimmed video. Existing fully-supervised TAL methods \cite{zhang2022actionformer,lin2021learning,shi2023tridet,weng2022efficient} have achieved promising performance by utilizing a large amount of labeled data. However, manually annotating temporal boundaries and class labels for large-scale datasets is very time-consuming and expensive. To address this issue, semi-supervised TAL (SS-TAL) methods \cite{ji2019learning,nag2022semi} have been proposed, where a large number of unlabeled videos and only a few labeled videos are leveraged for model training. 

One of SS-TAL frameworks \cite{ji2019learning,wang2021self,shi2021temporal} normally combines existing proposal-based TAL models with semi-supervised learning (SSL) approaches. However, this strategy suffers from the error propagation problem caused by sequential localization and classification design, resulting in accumulated errors. As an alternative, a proposal-free model \cite{nag2022semi} is proposed to address this problem, which designs a parallel localization and classification architecture. Despite the performance improvement, it fails to effectively suppress noisy pseudo labels. A recent work \cite{xia2023learning} elaborates on the importance of location biases and category errors, which focuses on alleviating the label noise problem. It combines adaptively learned class scores with subsequent manually calculated location scores (\ie, boundary variance) to measure the overall quality of pseudo labels but ignores the potential synergies or correlations between them due to its divergence from an end-to-end paradigm. Thus, how to effectively represent localization reliability remains an open question. More importantly, the above methods lack the ability to identify potential false positives and false negatives from pseudo labels selected based on fixed or
dynamical score thresholds, which hinders the full exploitation of positive instances generated on unlabeled data.

To address the above issues, in this paper, we propose a novel Adaptive Pseudo-label Learning (APL) framework for SS-TAL. Specifically, instead of manually computing the
location score based on the predicted boundaries, the proposed Adaptive Label Quality
Assessment (ALQA) jointly learns classification confidence and localization reliability of action instances by adopting an end-to-end learning architecture. For localization sub-task in most one-stage TAL methods \cite{zhang2022actionformer,shao2023action}, a Distance-IoU (DIoU) loss \cite{zheng2020distance} is normally performed to regress offsets from current frames to boundaries. Intuitively, the DIoU can evaluate the temporal intersection over union (overlap rate) between two segments (\ie, tIoU) and temporal normalized distance (tND) between predicted and ground-truth (GT) boundaries, which can be a natural indicator to assess localization quality. Motivated by this, we design two parallel branches to predict tIoU and tND, enabling the learning of localization reliability from different perspectives. The predictions are then dynamically divided into positives and candidates based on the joint score of classification and localization.     

Apart from the ALQA, we also propose an Instance-level Consistency Discriminator (ICD) to refine the selected pseudo labels on unlabeld videos by removing ambiguous positives (pseudo labels with high joint scores but wrong categories) and mining potential positives (pseudo labels with low joint scores but correct categories) from candidates. More concretely, the ICD is first trained using all labeled action instances to encourage feature consistency between different instances. Afterwards, during inference, each instance selected by the ALQA and the corresponding labeled instances having the same category are fed into the trained ICD to yield similarity probabilities, which 
serve as an auxiliary similarity score to further select high-quality pseudo labels.

 Self-supervised pre-training \cite{nag2022semi} has been investigated to improve the SS-TAL performance. However, this method relies on a customized and complex pretext task, which makes it difficult to be compatible with other backbones. More importantly, it aims to distinguish between actions and backgrounds but ignores discrimination between different actions. To this end, we design a general unsupervised ACP to enhance representation learning, which consists of coarse- and fine-grained contrasts. The former is implemented by contrasting binary-class frames (0 and 1 indicate foreground (\ie, action) and background in an untrimmed video) sampled from each video of a mini-batch, while the latter is performed by elaborately contrasting multi-class frames sampled from all videos of a mini-batch. In summary, the main contributions are as follows:
\begin{itemize}
\item We propose a SS-TAL framework APL, in which high-quality pseudo labels are adaptively selected to boost semi-supervised learning. Extensive experiments demonstrate that APL surpasses all the previous methods and achieves state-of-the-art performance.

\item We propose an ALQA module to facilitate more direct interaction between classification and localization via a joint learning paradigm, ensuring a robust quality assessment of pseudo labels.

\item We design an ICD for pseudo-label refinement, which aims to eliminate ambiguous positives and mine potential positives by leveraging the inherent consistency between distinct action instances.

\item We introduce a unsupervised ACP to enhance frame-level representation and improve the discrimination both within actions and between actions and backgrounds.

\end{itemize}

\section{Related Works}

\textbf{Temporal Action Localization.}
TAL involves simultaneously localizing and identifying action instances from untrimmed videos. Similar to the development of object detection \cite{ren2015faster,redmon2016you,tian2019fcos}, existing fully-supervised approaches can be divided into two categories, namely, two-stage methods \cite{lin2019bmn,lin2018bsn,yang2022temporal,qing2021temporal,su2023multi,bai2020boundary} and one-stage methods \cite{shi2022react,zhang2022actionformer,shi2023tridet,shao2023action,li2017joint}. Two-stage methods in TAL typically follow a proposal-generation and action-classification paradigm. Previous two-stage methods \cite{escorcia2016daps,lin2018bsn,lin2019bmn,zhao2020bottom,lin2020fast} usually focused on action proposal generation, where anchor-based works \cite{lin2020fast,escorcia2016daps,gao2018ctap} classify actions from specific anchor windows while boundary-based methods \cite{zhao2020bottom,lin2019bmn,lin2018bsn} predict the boundary probability and apply boundary matching mechanism to produce candidate proposals. Recent efforts have aimed at refining proposals by exploring temporal correlations between them using graph networks \cite{zeng2019graph,zhao2021video} or self-attention mechanisms \cite{zhu2021enriching,qing2021temporal}. Two-stage pipeline has shown effectiveness in handling complex temporal structures but may exhibit limitations in terms of computational efficiency. The one-stage methods integrate action localization and classification into a single network without using action proposals. Most previous works \cite{lin2017single,lin2021learning} utilized convolutional neural networks (CNNs) for feature encoding. Inspired by the recent success of DETR \cite{carion2020end}, transformer-based models \cite{zhang2022actionformer,shi2022react,shi2023tridet} have been designed for localization and classification, achieving new state-of-the-art performance. 

\noindent\textbf{Semi-Supervised Learning.} SSL that aims to improve model generalization and performance using a small amount of labeled data and a large amount of unlabeled data has been widely applied in various computer vision tasks, such as image classification \cite{tarvainen2017mean,sohn2020fixmatch}, action recognition or segmentation \cite{singh2021semi,xing2023svformer,zhou2023smc}, object detection \cite{chen2022dense,liu2023mixteacher} and semantic segmentation \cite{ouali2020semi,yang2023revisiting}. Consistency regularization \cite{laine2016temporal,miyato2018virtual,tang2021humble} and pseudo-labeling \cite{lee2013pseudo,sohn2020fixmatch} are two main paradigms in SSL. Consistency regularization methods, such as Mixmatch \cite{berthelot2019mixmatch}, aim to enforce consistency between predictions made on different perturbations of the same input. Pseudo-labeling methods exploit unlabeled data by training on self-generated predictions, \ie, pseudo labels. However, they are susceptible to low-quality pseudo labels due to inaccuracies in the model's predictions, leading to incorrect labels being assigned to the unlabeled data. As a result, another line of work \cite{wang2018iterative,yang2022class,li2023semi,chen2022label,chen2022dense,liu2023ambiguity} has attempted to tackle label noise. Our method addresses this issue in SS-TAL with a novel framework, where ALQA jointly learns classification confidence and localization reliability to better evaluate pseudo-label quality, while ICD eliminates ambiguous positives and mines potential positives to refine pseudo labels. 

\noindent\textbf{Semi-Supervised Temporal Action Localization.} Despite promising results in TAL, SS-TAL still remains insufficiently explored. A common pipeline of existing methods \cite{ji2019learning,wang2021self,shi2021temporal} is to incorporate SSL methods into fully-supervised TAL models. However, the accumulated errors caused by localization error propagation are inevitable. SPOT \cite{nag2022semi} addresses this problem with a proposal-free framework where localization and classification heads are constructed in a parallel manner. A very recent study, \ie, NPL \cite{xia2023learning}, aims to handle noisy pseudo labels by reranking the predictions according to the learned classification score and manually computed localization score.

\section{Proposed Method}

\textbf{Problem Definition.} In SS-TAL,  we aim to train a model using a small amount of labeled videos $\{V_{i}\}_{i=1}^{N_{l}} 
$ and a large amount of unlabeled videos $\{U_{i}\}_{i=1}^{N_{u}} 
$, where $N_{l}$ and $N_{u}$ indicate the numbers of labeled and unlabeled videos, respectively. Each labeled video $V_{i}$ is composed of a set of action instances  
$\mathcal{I}_{i} =\left \{{I}_{j}=(X_{j}, (t_{s,j},t_{e,j},y_{j})) \right \}_{j=1}^{M_{i}} 
$, where $M_{i}$ is the number of action instances,  and $t_{s,j}$, $t_{e,j}$ are the starting and ending time of the $j$-th action instance ${I}_{j}$. $y_{j}\in Y=\left \{ 0,1,...,K-1 \right \}$ is the class label, where $K$ denotes the number of classes. $X_{j}$ represents the features of ${I}_{j}$. Following the common practice in previous work \cite{zhang2022actionformer,xia2023learning}, we adopt video features extracted by a pre-trained video encoder (\eg, I3D network \cite{carreira2017quo}) as the input of our model, and sample a fixed number of frames for training. Therefore, each video $V_{i}$ can be represented as $V_{i}\in \mathbb{R}^{D \times T}$, where $D$ is the feature dimension and $T$ is the total number of frames. 

\noindent  \textbf{Approach Overview.}  Following the recent work \cite{xia2023learning} for SS-TAL, we leverage a one-sage detector, \ie, ActionFormer \cite{zhang2022actionformer} as our baseline, where each video frame is directly supervised by the corresponding labels, including distances to action boundaries and the action category.

\begin{figure*}[tp]
\begin{center}
\includegraphics[width=10.5cm]{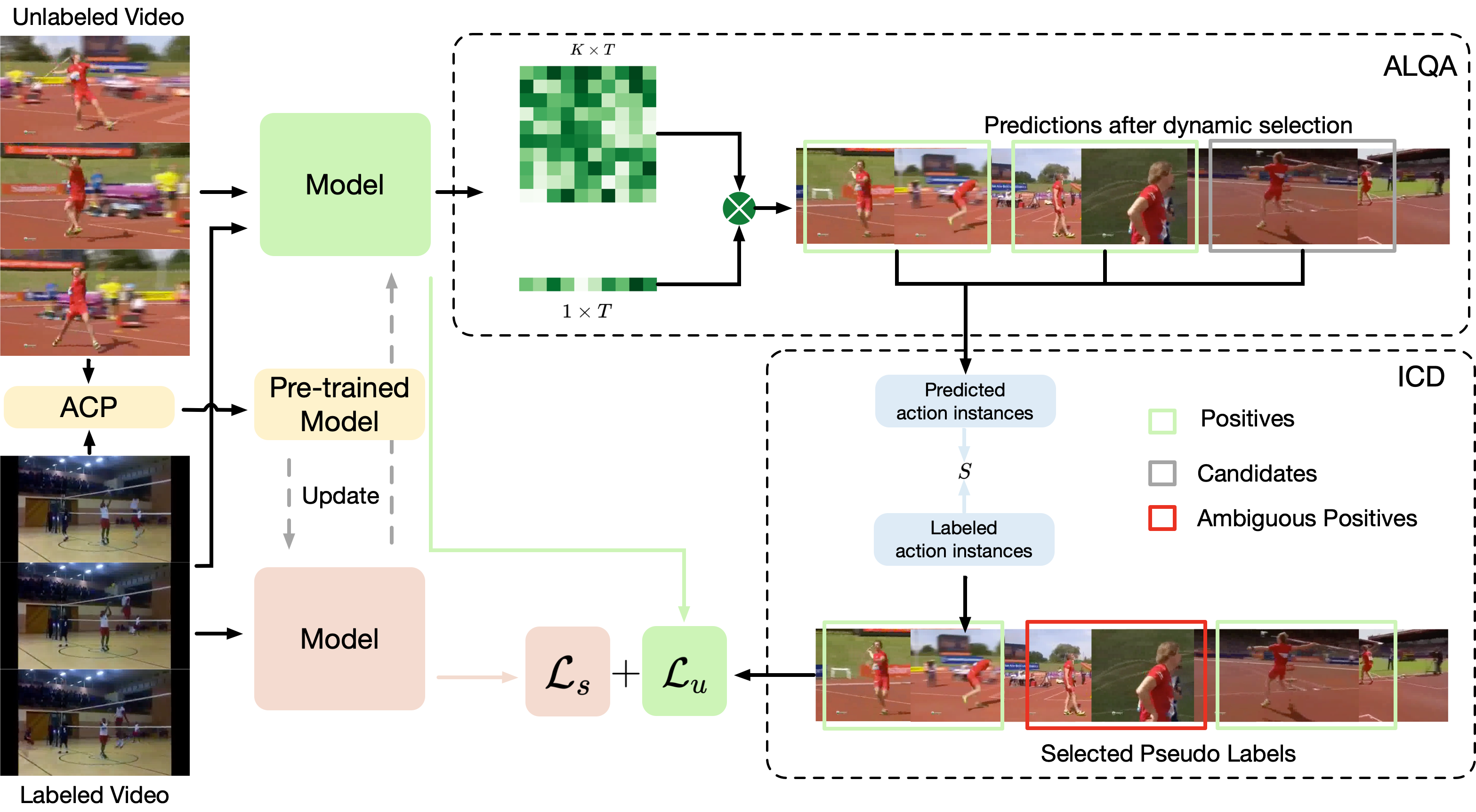}
\end{center}
\caption{The overview of the proposed APL framework. We first leverage both labeled and unlabeled videos for ACP without using any GT labels, which enhances the frame-level representation. We then update the pre-trained model by using a small amount of labeled videos and generate pseudo labels for unlabeled videos, where ALQA jointly learns classification confidence $\hat{P}_{cls}\in  \mathbb{R}^{K \times T}$ and localization reliability $\hat{P}_{diou}\in  \mathbb{R}^{1 \times T}$ before dynamically selecting pseudo labels according to their joint score. Finally, we propose an ICD to refine the pseudo-label selection by removing ambiguous positives and mining potential positives. 
}
\label{fig:framework}
\end{figure*}

In this paper, we propose a novel architecture, named APL, to adaptively explore high-quality pseudo labels, as shown in Fig. \ref{fig:framework}. Specifically, the proposed ALQA jointly learns classification confidence and
localization reliability in a fully adaptive fashion, thus achieving better ranking and selection of pseudo labels (Sec. \ref{sec:3.1}). Besides, the ICD aims to further refine the pseudo-label selection by discovering false positives and false negatives (Sec. \ref{sec:3.2}). Finally, the ACP is also introduced to generate more discriminative frame-level representation (Sec. \ref{sec:3.3}).

\subsection{Adaptive Label Quality Assessment}
\label{sec:3.1}

\begin{figure*}[tp]
\begin{center}
\includegraphics[width=10.5cm]{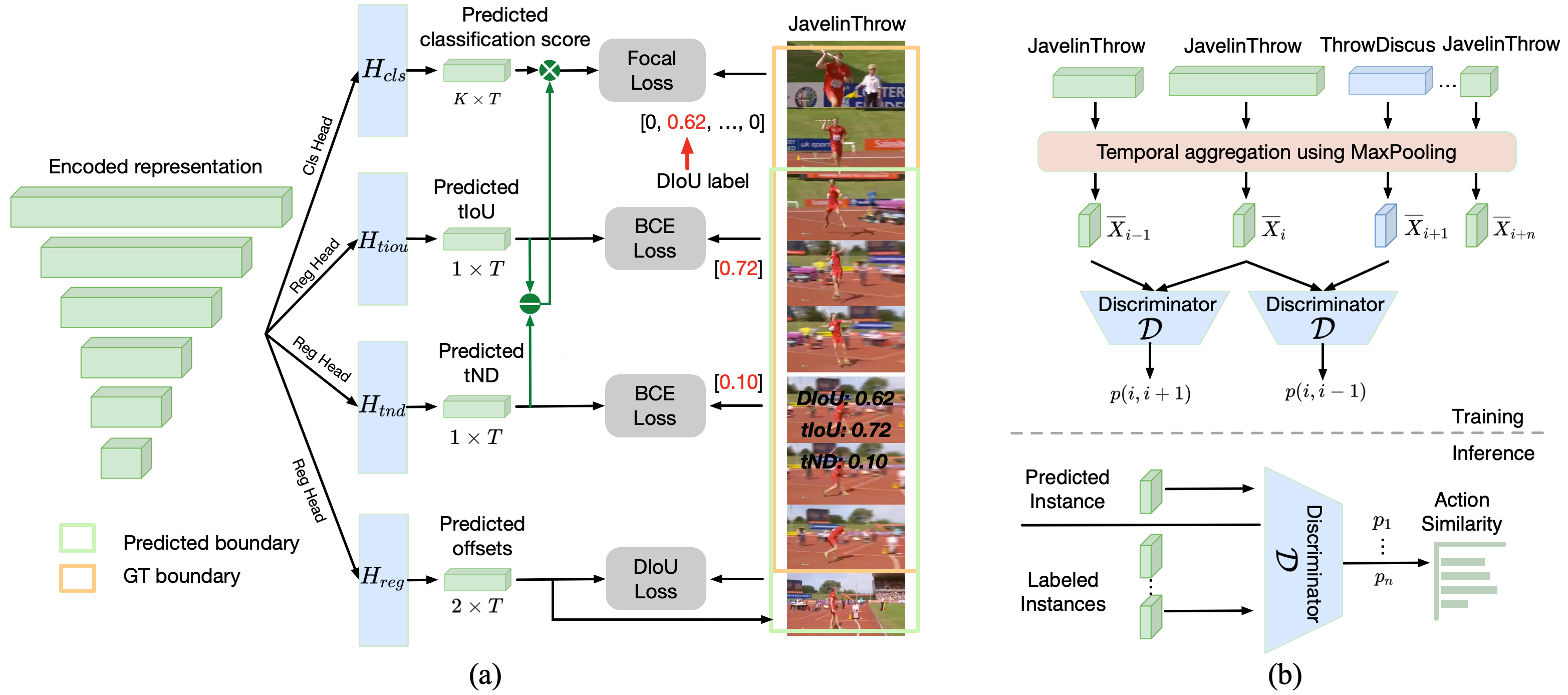}
\end{center}
\caption{Illustration of (a) Adaptive Label Quality Assessment and (b) Instance-level Consistency Discrimination. (a) We evaluate localization reliability by designing two parallel branches (heads) to predict tIoU and tND, respectively, leading to a joint score of classification and localization. (b) We aggregate temporal features of labeled action instances using Maxpooling, then use a discriminator $\mathcal{D}$ to learn the similarity probability between two instance pairs. During inference, $\mathcal{D}$ provides similarity scores between predicted instances and labeled instances of the same action category.}
\label{fig:ALQA_ICD}
\end{figure*}

In SS-TAL, the accurate assessment of classification and localization quality of pseudo labels is paramount. Recently, the determination of localization scores has relied on manual computations based on predicted boundaries \cite{xia2023learning}. In contrast, our proposed method, ALQA, introduces a novel paradigm by jointly learning the classification confidence and localization reliability of action instances, as shown in Fig. \ref{fig:ALQA_ICD}(a).  By jointly learning classification confidence and localization reliability, ALQA provides a unified framework for assessing the quality of pseudo labels. This allows for more accurate and reliable evaluation, as it considers the interaction between classification and localization.

As aforementioned, the regression head of most TAL methods, \eg, ActionFormer \cite{zhang2022actionformer}, involves the use of the DIoU loss for learning action boundaries. We leverage the intuitive qualities of DIoU, which evaluates both tIoU and tND between predicted and GT boundaries. Based on the original DIoU loss \cite{zheng2020distance} for bounding box regression, the DIoU loss used for TAL \cite{zhang2022actionformer} is formulated as:
\begin{equation}
\begin{split}
DIoU = tIoU-tND
= tIoU-\frac{\rho^{2}(c_{pre},c_{gt}) }{d^{2}}
\label{equation:diou}
\end{split}
\end{equation} 
where $d$ is the length of the smallest temporal box covering the predicted and GT boundaries and $\rho(c_{pre},c_{gt})$ is the distance between central points of the predicted and GT boundaries.

Based on the above observation, we design two parallel branches that can be seamlessly integrated into the original regression head of the baseline. They are dedicated to predicting tIoU and tND, respectively, offering distinct perspectives to learn localization reliability. Formally, given the encoded multi-scale representation $F$ (here we take the scale with $T$ frames for example), the predicted tIoU ($\hat{P}_{tiou}$) and tND ($\hat{P}_{tnd}$) can be formulated as:
\begin{equation}  
\begin{split}
\hat{P}_{tiou}=\operatorname{sigmod}(H_{tiou}(F))\in  \mathbb{R}^{1 \times T},
\hat{P}_{tnd}=\operatorname{sigmoid}(H_{tnd}(F))\in  \mathbb{R}^{1 \times T}
\label{equation:Hiou}
\end{split}
\end{equation}
where $H_{tiou}$ and $H_{tnd}$ have a similar structure as the regression head $H_{reg}$ of the baseline, consisting of 1D convolutions and layer normalization. $H_{tiou}$, $H_{tnd}$ and $H_{reg}$ share the same parameters, except for the final layer.

In the baseline, the original regression head aims to predict offsets from current frames to predicted boundaries, and the DIoU loss defined in Eq. (\ref{equation:diou}) is used to minimize the tIoU and tND between predictions and ground truth. In this way, we take the tIoU (${P}_{tiou}$) and tND (${P}_{tnd}$) calculated in this branch as the ground truth of our proposed tIoU and tND branches, respectively, and the overall loss of localization quality can be defined as follows:
\begin{equation}
\begin{split}
\mathcal{L}_{locq}=\operatorname{BCE}\left(\hat{P}_{tiou}, {P}_{tiou}\right)+\operatorname{BCE}\left(\hat{P}_{tnd}, {P}_{tnd}\right)
\label{equation:iou_nd}
\end{split}
\end{equation}
where BCE denotes the binary cross entropy. The outputs, \ie, $\hat{P}_{tiou}$ and $\hat{P}_{tnd}$, are probability values between 0 and 1. Thus, we use BCE loss to measure the difference between the model's output and the ground truth, guiding the model optimization. For the classification branch, the joint score $\hat P$ (between 0 and 1) can be obtained by combining the classification score $\hat{P}_{cls}$ predicted by the original classification head and the localization reliability consisting of the $\hat{P}_{tiou}$ and $\hat{P}_{tnd}$, defined as:
\begin{equation}  
\begin{split}
\hat P =\hat{P}_{diou} \odot \hat{P}_{cls}=\max[\hat{P}_{tiou}-\hat{P}_{tnd},\epsilon]\odot \hat{P}_{cls}
\label{equation:jointscore}
\end{split}
\end{equation}
where $\odot$ denotes element-wise product. The difference $\hat{P}_{diou}$ between $\hat{P}_{tiou}$ and $\hat{P}_{tnd}$ represents the overall localization reliability. Particularly, to avoid negative reliability probabilities during training, we set the difference to a small positive value  $\epsilon$ when it is negative. This is because a negative difference indicates a relatively large temporal distance between the predictions and the ground truth, representing a low reliability probability.

Similar to the baseline, the objective function in the classification branch is also based on focal loss (FL) \cite{zhang2022actionformer,shi2022react}. However, the classification target of these methods normally uses one-hot label encoding, which is less consistent with the predicted joint score. Inspired by the soft label mechanism \cite{li2020generalized,liu2023ambiguity}, we replace the one-hot label with a combination of predicted tIoU, tND and the one-hot label to facilitate unified optimization, promoting direct interaction between classification and localization.
Thus, we form a DIoU-based soft label $P=\left \{ 0,\dots , {P}_{diou},\dots ,0  \right \}$, where ${P}_{diou}={P}_{tiou}-{P}_{tnd}$ (we also set ${P}_{diou}$ to a small positive value during training when it is negative). The overall focal loss combining classification and localization qualities is formulated as: 
\begin{equation}  
\begin{split}
\mathcal{L}_{cls}=\operatorname{FL}\left(\hat{P}, {P}\right)
\label{equation:focal}
\end{split}
\end{equation}

After generating predictions on unlabeled data, we apply Soft-NMS \cite{bodla2017soft} to remove redundant and low-quality instances. Unlike previous methods that rely on fixed confidence thresholds or frequency \cite{xia2023learning} of different actions in pseudo labels to select positive instances, our approach introduces a selection mechanism with adaptive criteria.  Specifically, given the sets of predicted instances $\hat {\mathcal{I}}_{u}= \{ \hat I_{i} \}_{i=1}^{L_{nms}}$ and corresponding joint scores $\hat {\mathcal{P}}_{u}= \{ \hat P_{i} \}_{i=1}^{L_{nms}}$ where $L_{nms}$ is the number of instances after NMS, we construct positives $\hat {\mathcal{I}}_{pos}$ and candidates $\hat {\mathcal{I}}_{can}$ based on the thresholds of joint scores:
\begin{equation}
\begin{split}
  \hat {\mathcal{I}}_{pos}=\{ \hat I_{i}\mid \hat P_{i}>=\tau_{pos} \}_{i=1}^{L_{nms}},  \hat {\mathcal{I}}_{can}=\{ \hat I_{i}\mid \hat \tau_{neg}<\hat P_{i}<\tau_{pos} \}_{i=1}^{L_{nms}}
\label{equation:split_instance}
\end{split}
\end{equation}
where $\tau_{neg}$ is fixed to 0.15 to directly remove low-quality instances. $\tau_{pos}$ is dynamically computed based on the mean and standard deviation of joint scores $\{\hat P_{i}\mid\hat P_{i}>\tau_{neg} \}_{i=1}^{L_{nms}} $ of the remaining instances.

\subsection{Instance-level Consistency Discrimination}
\label{sec:3.2}
Although ALQA can achieve better ranking and selection of pseudo labels, there still exist some ambiguous positives and potential positives in $\hat{\mathcal{I}}_{pos}$ (defined in Eq. (\ref{equation:split_instance})) and  $\hat{\mathcal{I}}_{can}$, respectively. To tackle this problem, we propose the ICD to learn and leverage inter-instance intrinsic consistency,  ensuring a more comprehensive and accurate identification of positive instances within the predictions.

As shown in Fig. \ref{fig:ALQA_ICD}(b), during training on labeled videos, we sample the $i$-th instance of one mini-batch and the corresponding input feature and action category are defined as  $X_{i}\in \mathcal{X}_{i}^{b}=\{X_{i}\}_{i=1}^{ M_{b}}$ and $y_{i}$, respectively, where $M_{b}$ is the total number of action instances within a mini-batch. Two feature sets are then constructed as: 
\begin{equation}  
\begin{split}
\mathcal{G}_{i}=\{X_{r}|y_{r}=  y_{i}, 1\le r\le M_{b} \},
\overline{\mathcal{G}}_{i}=\{X_{q}|y_{q}\ne   y_{i}, 1\le q\le  M_{b} \}
\label{equation:ICD_set}
\end{split}
\end{equation}
where $\mathcal{G}_{i}$ comprises features of instances sharing the same action category as the $i$-th instance, whereas $\overline{\mathcal{G}}_{i}$ encompasses features of instances with distinct action categories. Notably, in our implementation, we sample a fixed number of instances for the two sets to ensure a balanced training in Eq. (\ref{equation:ICD_loss}). 

Subsequently, the features (\eg, $X_{i}$ $vs$ $X_{r}$)  of instances with the same action type should be similar, while those (\eg, $X_{i}$ $vs$ $X_{q}$ ) with different action types should be dissimilar. Therefore, the overall objective function of the ICD is formulated as follows:
\begin{equation} 
\begin{split}
\mathcal{L}_{icd} =-\mathbb{E}_{X_{i}\sim\mathcal{X}_{i}}\big[\mathbb{E}_{X_{r}\sim\mathcal{G}_{i}}[\log\mathcal{D}(\overline X_{i},\overline X_{r})]+\mathbb{E}_{X_{q}\sim\overline{\mathcal{G}}_{i}}[\log(1- \mathcal{D}(\overline X_{i},\overline X_{q}))]\big]
\label{equation:ICD_loss}
\end{split}
\end{equation}
where $\mathbb{E}$ denotes the expectation. $\overline X_{i}=\operatorname{MAP}(X_i)
$ and $\operatorname{MAP}(\cdot)$ is the MaxPooling operation that aggregates temporal features of action instances. $\mathcal{D}$ is the discriminator that aims to predict the probability of pair-wise instances being similar,  which works as follows:
\begin{equation}  
\begin{split}
\mathcal{D}(\overline X_i,\overline X_{r})= \operatorname{MLP} ([\overline X_i; \overline X_{r}]),\mathcal{D}(\overline X_i,\overline X_{q})= \operatorname{MLP} ([\overline X_i; \overline X_{q}])
\label{equation:ICD_prob}
\end{split}
\end{equation}
where $[;]$ represents the concatenation operation along the feature dimension. $\operatorname{MLP}$ maps the input features from $2D$ to 1. The objective is to ensure that the output probability of $\mathcal{D}(\overline X_i, \overline X_{r})$ approaches 1, while that of $\mathcal{D}(\overline X_i, \overline X_{q})$ tends toward 0. Consequently, we leverage the BCE loss for the optimization of our ICD, which is independent of the baseline’s training procedure. 

In the inference phase, the ICD is employed to produce similarity scores, reflecting the overall similarity between a predicted instance and the labeled instances belonging to the same category. 
Mathematically, when considering the action category $\hat y_{i}$ and the corresponding input feature $\hat X_i$ of a predicted instance, we begin by selecting all GT instances with the same category from labeled videos. The feature set of these instances, denoted as $\mathcal{X}_{i}^{l}$, can be expressed as $\mathcal{X}_{i}^{l} =\{X_{j}|y_{j}=\hat y_{i}  \}_{j=1}^{M_{l}}$, 
where $M_{l}$ denotes the total number of instances, and $y_{j}$ represents the action class of the $j$-th instance. We then pass the predicted instance with feature $\hat X_i$ and each instance from $\mathcal{X}_{i}^{l}$ to the ICD, resulting in the average similarity score: 
\begin{equation} 
\begin{split}
S_{i}=\frac{1}{|\mathcal{X}_{i}^{l}|} \sum_{X_{j}\in \mathcal{X}_{i}^{l}}^{} \mathcal{D}(\operatorname{MAP}(\hat X_i),\operatorname{MAP}(X_{j}))
\label{equation:ICD_infer2}
\end{split}
\end{equation}

The computed similarity score helps identify ambiguous positives from $\hat {\mathcal{I}}_{pos}$ and reveal potential positives from $\hat {\mathcal{I}}_{can}$. Since the overall quality of pseudo labels from $\hat {\mathcal{I}}_{can}$ is lower than those from $\hat {\mathcal{I}}_{pos}$, we set a relatively high threshold $\varsigma_{icd}$ ($>$ 0.5) to select positive instances from $\hat {\mathcal{I}}_{can}$. To retain more positives from $\hat {\mathcal{I}}_{pos}$, instances with a similarity score below $\tau_{icd}$ are excluded (see Fig. \ref{fig:parameter_curve}).

\subsection{Action-aware Contrastive Pre-training}
\label{sec:3.3}
While the self-supervised pre-training \cite{nag2022semi} has demonstrated efficacy for SS-TAL, the reliance on intricate pretext tasks and neglect of distinctions between actions limit its application and performance. The proposed ACP seeks to provide a general and unsupervised alternative to enhance frame-level representation, which can be combined with various backbone models.

The ACP involves two types of contrasts, \ie, coarse- and fine-grained contrasts (Fig. S1 in \textbf{Supp. C}). The former aims at contrasting pair-wise frames within an untrimmed video, with a specific emphasis on distinguishing actions and backgrounds. Our ACP is performed on the multi-scale representation encoded by the FPN neck of ActionFormer \cite{zhang2022actionformer}. To implement the coarse-grained contrast, we first upsample each representation to match the temporal length of the input and then concatenate these representations along the feature dimension to generate a new representation. We then partition the representation of each video in a mini-bath into $N$ equal segments. Subsequently, a single frame is randomly selected from each partition, forming the representation set of each video as the input for our ACP. Since the ACP follows a unsupervised setting, no GT action labels are provided to guide contrastive learning. Hence we perform K-means clustering on the corresponding input features to generate initial action classes and the number of clusters is set to 2. Formally, let $f_{i} \in \mathcal{F}=\{f_{i} \}_{i=1}^{N}$ and  $l_{i} \in \{0,1\}$ denote the $i$-th feature of the representation set $\mathcal{F}$ and the corresponding clustering labels, the positive sets $\mathcal{P}_{i}$ and negative sets $\mathcal{N}_{i}$ are constructed as $\mathcal{P}_{i}=\{f_{j}|l_{j}=l_{i} \}_{i=1}^{N}$ and $\mathcal{N}_{i}=\{f_{j}|l_{j}\ne l_{i} \}_{i=1}^{N}$, respectively.
We then use infoNCE loss \cite{oord2018representation} for coarse-grained contrast, which is defined as follows:
\begin{equation}  \small
\begin{split}
\mathcal{L}_{conc}=-\frac{1}{N_{c}}\sum_{f_{i}}^{}\sum_{f_{j}\in \mathcal{P}_{i}}^{} \operatorname{log} \frac{\operatorname{exp}(\operatorname{sim}(f_{i},f_{j})/\varsigma )}{\operatorname{exp}(\operatorname{sim}(f_{i},f_{j})/\varsigma )+\sum_{f_{*}\in \mathcal{N}_{i}}^{}\operatorname{exp}(\operatorname{sim}(f_{i},f_{*})/\varsigma ) } 
\label{equation:loss_conc}
\end{split}
\end{equation}
where $N_{c}= {\textstyle \sum_{i}^{}} | \mathcal{P}_{i}|$, $\operatorname{sim}(\cdot)$ is the inner product between two normalized vectors and $\varsigma >0$ is a temperature parameter.

Complementing the coarse-grained contrast, the fine-grained contrast aims to improve the discrimination between actions by elaborately contrasting frames from all videos in a mini-batch. To achieve this, we combine the representation sets of all videos and obtain the fine-grained clustering labels. Different from binary-class labels used in coarse-grained contrast, we assign multi-class labels to the frames of the combined set, where $l_{i} \in \{0,1,...,B-1\}$. Here, $B$ is the number of clusters and is determined based on the batch size and datasets. Similarly, the fine-grained contrast loss $\mathcal{L}_{conf}$ can be computed using Eq. (\ref{equation:loss_conc}), and the overall pre-trained loss is represented as $\mathcal{L}_{acp}= \mathcal{L}_{conc}+\mathcal{L}_{conf}$. Note that $\mathcal{L}_{acp}$ is also employed for fine-tuning on labeled videos with GT action classes and unlabeled videos with pseudo labels after pre-training (see Tab. S4).

Finally, the overall objective loss for our SS-TAL framework is designed as:
\begin{equation} 
\begin{split}
\mathcal{L}&=\mathcal{L}^{s}+\beta \mathcal{L}^{u}\\
&=\frac{1}{N_{pos}^{s}}\sum_{t}^{}(\mathcal{L}_{cls}^{s}+\lambda_{reg}\mathbbm{1}_{in_t}\mathcal{L}_{reg}^{s}+\lambda_{locq}\mathbbm{1}_{in_t}\mathcal{L}_{locq}^{s})+\lambda_{acp}\mathcal{L}_{acp}^{s}+\mathcal{L}_{icd}\\  
&+\beta (\frac{1}{N_{pos}^{u}  }\sum_{t}^{}(\mathcal{L}_{cls}^{u}+\lambda_{reg}\mathbbm{1}_{in_t}\mathcal{L}_{reg}^{u}+\lambda_{locq}\mathbbm{1}_{in_t}\mathcal{L}_{locq}^{u})+\lambda_{acp}\mathcal{L}_{acp}^{u})
\label{equation:total_loss}
\end{split}
\end{equation}
where $\beta$ is the weight of unsupervised loss, which is set to 2. $\mathbbm{1}_{in_t}$ is an indicator denoting whether the $t$-th frame is within a GT action or background. $N_{pos}^{*}$ is the number of frames within action segments. $\mathcal{L}_{reg}^{*}$ indicates the DIoU loss and $\lambda_{reg}$ is set to the default value in \cite{zhang2022actionformer}. Both $\lambda_{locq}$ and $\lambda_{acp}$ are set to 0.1.

\section{Experiments}
\label{sec:experiments}

\noindent \textbf{Datasets and Evaluation.} We evaluate our method on two challenging TAL benchmarks, \ie, THUMOS14 \cite{THUMOS14} and ActivityNet v1.3 \cite{caba2015activitynet}. We report the mean average precision (mAP) at different tIoU thresholds. The thresholds are [0.3:0.1:0.7] for THUMOS14 and [0.5:0.05:0.95] for ActivityNet v1.3. Following \cite{xia2023learning}, we randomly select 10\%, 20\%, 40\% and 60\% of the training videos as labeled data and the remaining as unlabeled data.

\noindent \textbf{Implementation Details.} Similar to \cite{xia2023learning}, our SS-TAL framework is also based on the detector ActionFormer \cite{zhang2022actionformer}. In addition, we combine the proposed components with BMN \cite{lin2019bmn}, which is a two-stage proposal-based detector. For fair comparisons, we use two popular backbones, \ie, TSN \cite{wang2016temporal} and I3D \cite{carreira2017quo} pre-trained on Kinetics to extract the video features. More implementation details of our semi-supervised learning are provided in the supplementary material.

\begin{table*}[ht]
\centering
\caption{Comparison with the state-of-the-art methods on THUMOS14 and ActivityNet v1.3. We report mAP ($\%$) at different tIoU thresholds. ActF refers to ActionFormer \cite{zhang2022actionformer}. * means using only labeled training videos.}
\label{tab:comparison}
\resizebox{0.95\textwidth}{5.5cm}{\begin{tabular}
{c|c|c|ccccc|c|ccc|c}
\toprule
\multirow{2}{*}{Label} & \multirow{2}{*}{Method} & \multirow{2}{*}{Backbone} & \multicolumn{6}{c|}{THUMOS14}& \multicolumn{4}{c}{ActivityNet v1.3}  
\\ \cmidrule{4-13}
 &  &  & 0.3 & 0.4 & 0.5 & 0.6 & 0.7 & Avg. & 0.5 & 0.75 & 0.95 &  Avg. \\ \midrule
\multirow{9}{*}{ 10\%} & ActF* \cite{zhang2022actionformer} & I3D & 28.5 & 22.9 & 14.1 & 8.2 & 4.1 & 15.6 & 47.8 & 24.2 & 1.7 &  25.6 \\
 & ActF + MixUp \cite{zhang2017mixup} & I3D & 29.7 & 24.2 & 14.5 & 9.6 & 5.4 & 16.7 & 49.4 & 27.9 & 3.1 &  28.8 \\
 & NPL (ActF) \cite{xia2023learning} & I3D & 32.8 & 29.6 & 20.1 & 11.7 & 7.2 & 20.3 & 51.9 & 33.4 & 3.6 &  32.5 \\
 & APL (ActF) & I3D & \textbf{35.1} & \textbf{31.7} & \textbf{25.6} & \textbf{19.1} & \textbf{11.0} & \textbf{24.5} &\textbf{52.2} &\textbf{33.9} &\textbf{6.7} &\textbf{33.5}\\
\cmidrule{2-13}
& SSP \cite{ji2019learning}  & TSN & 44.2 & 34.1 & 24.6 & 16.9 & 9.3 & 25.8 & 38.9 & 28.7 & 8.4 & 27.6 \\
& SSTAP \cite{wang2021self}   & TSN & 45.6 & 35.2 & 26.3 & 17.5 & 10.7 & 27.0 & 40.7 & 29.6 & \textbf{9.0}& 28.2 \\
 & SPOT \cite{nag2022semi} & TSN & 49.4 & 40.4 & 31.5 & 22.9 & 12.4 & 31.3 & 49.9 & 31.1 & 8.3 & 32.1 \\
 & NPL (BMN)  \cite{xia2023learning} & TSN & 50.0 & 41.7 & 33.5 & 23.6 & 13.4 & 32.4 & 50.9 & 32.0 & 7.9 & 32.6 \\
 & APL (BMN)  & TSN & \textbf{51.5}  & \textbf{42.5} &\textbf{34.6} & \textbf{24.4} & \textbf{13.5} & \textbf{33.3} & \textbf{51.5} & \textbf{32.4} & 8.2 &\textbf{33.0}   \\
 \midrule
\multirow{7}{*}{ 20\%} & ActF* \cite{zhang2022actionformer} & I3D & 49.1 & 41.6 & 32.6 & 21.5 & 12.1 & 31.4 & 51.2 & 34.3 & 3.8 &  32.9 \\
 & ActF + MixUp \cite{zhang2017mixup}  & I3D & 51.2 & 43.2 & 34.0 & 23.9 & 14.1 & 33.3 & 52.9 & 34.7 & 3.9 &  33.3 \\
 & NPL (ActF) \cite{xia2023learning} & I3D & 54.5 & 47.1 & 39.3 & 29.7 & 18.5 & 37.8 & 53.1 & 35.8 & 3.9 &  33.8 \\
 & APL (ActF)  & I3D & \textbf{59.2} & \textbf{54.2} & \textbf{44.5} & \textbf{34.2} & \textbf{22.1} & \textbf{42.8} &\textbf{53.5} &\textbf{36.1} &\textbf{7.1} &\textbf{34.5}\\
 \cmidrule{2-13}
 & SPOT \cite{nag2022semi}  & TSN & 52.6 & 43.9 & 34.1 & 25.2 & 16.2 & 34.4 & 51.7 & 32.0 & 6.9 & 32.3 \\
 & NPL (BMN)  \cite{xia2023learning} & TSN & 53.9 & 45.6 & 36.2 & 26.9 & 16.5 & 35.8 & 52.1 & 32.9 & 7.9 & 32.9 \\
 & APL (BMN)  & TSN  & \textbf{54.8}	&\textbf{45.9}	&\textbf{37.1}	&\textbf{28.5}	&\textbf{16.9	}&\textbf{36.6}	&\textbf{52.4}	&\textbf{33.3}	&\textbf{8.3}	&\textbf{33.4} \\
 \midrule
\multirow{7}{*}{ 40\%} & ActF* \cite{zhang2022actionformer} & I3D & 69.0 & 60.4 & 49.3 & 31.5 & 19.3 & 45.9 & 53.2 & 35.7 & 3.8 &  34.2 \\
 & ActF + MixUp \cite{zhang2017mixup} & I3D & 69.7 & 61.9 & 52.4 & 34.4 & 20.1 & 47.7 & 53.1 & 36.0 & 4.3 &  34.5 \\
 & NPL (ActF) \cite{xia2023learning} & I3D & 71.9 & 65.4 & 55.7 & 40.9 & 23.4 & 51.5 & 53.6 & 36.5 & 4.6 &  35.3 \\
 & APL (ActF) & I3D & \textbf{73.2} & \textbf{68.2} & \textbf{59.1} & \textbf{44.9} & \textbf{28.7} & \textbf{54.8}  &\textbf{53.8} & \textbf{36.7} & \textbf{8.2} & \textbf{35.5}\\
\cmidrule{2-13}
 & SPOT \cite{nag2022semi}& TSN & 54.4 & 45.8 & 37.2 & 29.7 & 19.4 & 37.3 & 53.3 & 33.0 & 6.6 & 33.2 \\
 & NPL (BMN)  \cite{xia2023learning} & TSN & 56.2 & 46.7 & 38.8 & 30.3 & 19.5 & 38.3 & 53.4 & \textbf{33.9} & 8.1 & 33.8 \\
 & APL (BMN)  & TSN & \textbf{57.0}	&\textbf{47.1}	&\textbf{39.5}	&\textbf{32.7}	&\textbf{20.1}	&\textbf{39.3}	&\textbf{53.5}	&33.8	&\textbf{8.5}	&\textbf{34.1}   \\
 \midrule
\multirow{9}{*}{ 60\%} & ActF* \cite{zhang2022actionformer} & I3D & 71.5 & 65.6 & 59.9 & 47.3 & 32.7 & 55.4 & 53.9 & 36.1 & 5.7 &  35.0 \\
 & ActF + MixUp \cite{zhang2017mixup} & I3D & 72.2 & 67.5 & 61.2 & 48.7 & 34.0 & 56.7 & 54.1 & 36.4 & 5.7 & 35.2 \\
 & NPL (ActF) \cite{xia2023learning} & I3D & 74.5 & 69.9 & 62.8 & 51.1 & 36.6 & 59.0 & 54.3 & \textbf{36.7} & 6.5&  35.8 \\
 & APL (ActF) & I3D &\textbf{77.3} & \textbf{73.1} &\textbf{65.0} & \textbf{52.4} & \textbf{37.6} & \textbf{61.1} &\textbf{54.4} & \textbf{36.7} & \textbf{8.4} &\textbf{36.0} \\
 \cmidrule{2-13}
& SSP \cite{ji2019learning}  & TSN & 53.2 & 46.8 & 39.3 & 29.7 & 19.8 & 37.8 & 49.8 & 34.5 & 7.0 & 33.5 \\
& SSTAP \cite{wang2021self}  & TSN & 56.4 & 49.5 & 41.0 & 30.9 & 21.6 & 39.9 & 50.1 & 34.9 & 7.4 & 34.0 \\
 & SPOT \cite{nag2022semi} & TSN & 58.9 & 50.1 & 42.3 & 33.5 & 22.9 & 41.5 & 52.8 & 35.0 & 8.1 & 35.2 \\
 & NPL (BMN)  \cite{xia2023learning} & TSN & 59.0 & 51.4 & 42.9 & 34.3 & 23.3 & 42.2 & 53.9 & 35.8 & 8.5 & 35.7 \\
 & APL (BMN)   & TSN &  \textbf{59.7}	& \textbf{51.6}	& \textbf{43.2}	& \textbf{34.9}	& \textbf{23.6	}& \textbf{42.6}	& \textbf{54.2}	& \textbf{36.2}	& \textbf{8.6}	& \textbf{35.9}  \\
\bottomrule
\end{tabular}}
\end{table*}

\subsection{Comparison with State-of-the-art Methods}
\textbf{THUMOS14.} Following \cite{xia2023learning}, we combine our APL with ActionFormer \cite{zhang2022actionformer} and BMN \cite{lin2019bmn}. For ActionFormer (using I3D features), we add two regression heads to predict tIoU and tND, which measure the localization reliability of pseudo labels.  As shown in Tab. \ref{tab:comparison}, our APL achieves superior performance and suppresses the state-of-the-art methods in mAP at different thresholds, which demonstrates its effectiveness. In particular, APL achieves 42.8\% in the average mAP on THUMOS14 with 20\% labeled data, which outperforms NPL by a large margin, namely 5\% absolute improvement. For the two-stage detector, we incorporate the proposed components into BMN (using TSN features) to facilitate the selection of action proposals. The results in Tab.  \ref{tab:comparison} show that the two-stage detector can also benefit from the proposed APL. Notably, APL obtains almost 2\% average mAP improvement compared with the recent anchor-free SPOT \cite{nag2022semi} when using 10\%, 20\% and 40\% labeled data.
 
\noindent\textbf{ActivityNet v1.3.} For the ActivityNet v1.3 dataset, we also adopt the I3D and TSN as our backbone features. With I3D features, our method reaches an average mAP of 33.5\% with 10\% labeled data, outperforming the closest competitor NPL by 1\%. It is noteworthy that our APL significantly improves the mAP when the tIoU threshold is set to 0.95 (mAP@0.95), resulting in improvements of 3.1\%, 3.3\%, 3.6\% and 1.9\% under different semi-supervised settings. The results are still comparable when using TSN features. Our method receives the best mAP@0.5 and average mAP compared to other SS-TAL frameworks, which can be attributed to the improvement in the quality of pseudo labels.

\subsection{Ablation Study}

To further verify the efficacy of our contributions, we conduct comprehensive ablation studies on the THUMOS14 dataset, including each component of our method and the choice of hyper-parameters. ActionFormer \cite{zhang2022actionformer} based on I3D \cite{carreira2017quo} is used as the localization framework.

\noindent\textbf{Effectiveness of each component.} We demonstrate the effectiveness of three proposed components in APL, including ALQA, ICD, and ACP. In Tab. \ref{tab:ablation_comp}, the baseline denotes the case where the pseudo labels are filtered according to a fixed classification confidence threshold (\ie, 0.3). Under the 10\% labeling ratio, we can see that compared with the baseline, our ALQA brings about a 4.3\% absolute improvement in the average mAP, proving the effectiveness of the module by jointly learning classification and localization quality. After being equipped with ICD, it boosts the performance by 2\% of average mAP, which demonstrates the effectiveness of ICD in refining the selection of pseudo labels. When further applying our pre-training strategy, the performance is increased to 24.5\% mAP. This substantiates that our ACP effectively facilitates SS-TAL. We can observe a similar trend when utilizing 40\% labeled videos.

\begin{table}[tp]  
\begin{minipage}{0.5\linewidth}
\caption{Effectiveness of three main components on THUMOS14, using 10\% and 40\% labeled videos. '+' means training by the proposed method.}
\centering
\resizebox{0.9\textwidth}{1.8cm}{\begin{tabular}{c|c|ccc|c}
\toprule
\multirow{2}{*}{Label} & \multirow{2}{*}{Method} &  \multicolumn{4}{c}{mAP(\%)} 
\\ \cmidrule{3-6}
 &  & 0.3 & 0.5 & 0.7 & Avg. \\ \midrule
\multirow{4}{*}{ 10\%} & baseline & 26.6 & 17.8 & 6.1 & 17.0 \\
&+ALQA &31.9&22.9 & 8.1 &21.3\\
&+ICD  &33.6 & 24.6 & 10.2 & 23.3\\
&+ACP & \textbf{35.1} & \textbf{25.6} & \textbf{11.0 }& \textbf{24.5} \\
\midrule
\multirow{4}{*}{ 40\%} & baseline &65.9 & 51.7 & 22.8 & 48.8\\
&+ALQA &71.0 &56.7 &26.8 & 52.6\\
&+ICD  &72.0 & 58.1 & 27.7 & 53.9\\
&+ACP  &\textbf{73.2} &  \textbf{59.1} & \textbf{28.7} & \textbf{54.8}\\
\bottomrule
\end{tabular}}
\label{tab:ablation_comp}
\end{minipage}%
\hfill
\begin{minipage}{0.48\linewidth}
\caption{The effect of different localization reliability learning strategies on THUMOS14 with 10\% and 40\% labeled videos.}
\centering            
\resizebox{0.9\textwidth}{1.8cm}{\begin{tabular}{c|c|ccc|c}
\toprule
\multirow{2}{*}{Label} & \multirow{2}{*}{Method} &  \multicolumn{4}{c}{mAP(\%)} 
\\ \cmidrule{3-6}
 &  & 0.3 & 0.5 & 0.7 & Avg. \\ \midrule
\multirow{4}{*}{ 10\%} 
& tIoU (1b) & 28.9 & 20.7 & 7.6 & 19.4\\
&DIoU (1b) & 30.7 & 20.8 & 8.0 & 20.0 \\
\cmidrule{2-6}
&tIoU+tND (2b)  &\textbf{31.9} & \textbf{22.9} & \textbf{8.1} & \textbf{21.3}\\
\midrule
\multirow{4}{*}{ 40\%} 
& tIoU (1b) & 68.4 & 54.8 & 24.4 & 50.7\\
&DIoU (1b) & 70.3 & 55.4 & 26.0 & 51.8 \\
\cmidrule{2-6}
&tIoU+tND (2b) &\textbf{71.0} &\textbf{56.7} &\textbf{26.8} & \textbf{52.6}\\
\bottomrule
\end{tabular}}
\label{tab:ablation_branch}
\end{minipage}
\end{table}

\noindent\textbf{Ablation on ALQA.} In this section, we present the ablation results for different localization reliability learning strategies in Tab. \ref{tab:ablation_branch}. tIoU (1b) denotes designing a single branch (1b) to learn the tIoU. With 10\% labeled data, the performance is increased from 19.4\% mAP to 20\% mAP by replacing the tIoU prediction with DIoU (Eq. (\ref{equation:diou})) prediction (1b). This shows that temporal distance can assist in evaluating the localization quality. We also find that the average mAP is further improved by 1.3\% when we design two different branches (2b) for predicting tIoU and tND, respectively. The main reason can be that having two separate branches enables the model to adaptively adjust its focus on different aspects of action localization, resulting in improved performance.

\noindent\textbf{Ablation on ICD.} To study how ICD affect performance, we separately apply EAP, MPP and their combination (\ie, EAP + MPP) to optimize the pseudo labels. In Tab. \ref{tab:ICD}, we see that MPP plays a more important role in filtering pseudo labels than EAP and merging them can gain further promotion.

\begin{figure}[tp]
\begin{center}
\includegraphics[width=10cm]{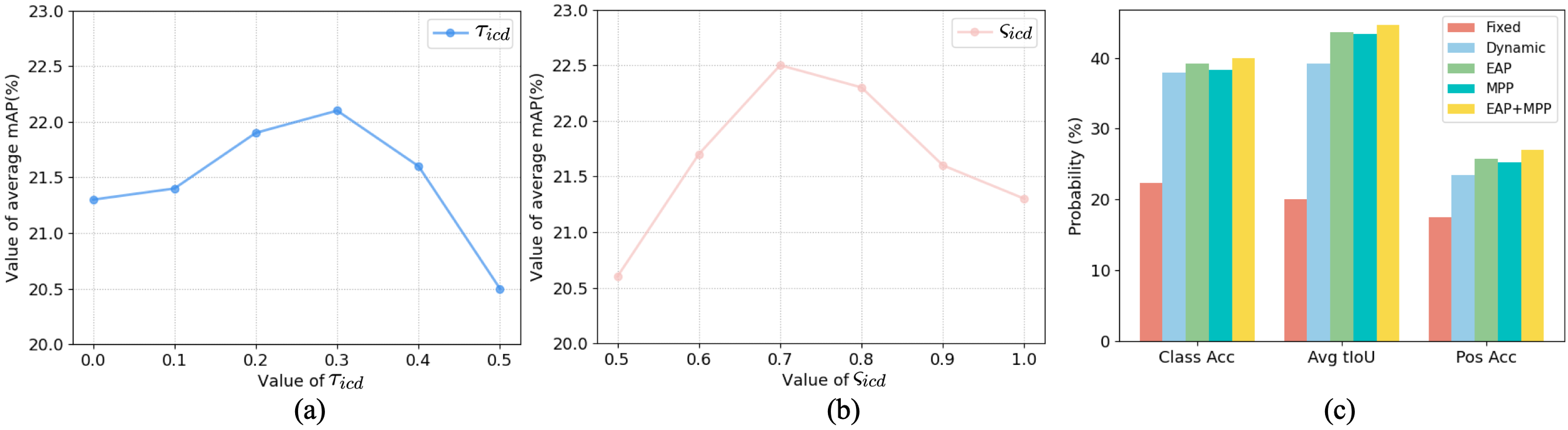}
\end{center}
\caption{(a) and (b) The effect of different hyperparameters (\ie, $\tau_{icd}$ and $\varsigma_{icd}$) settings. (c) Ablation studies on the quality of pseudo labels when using 10\% labeled videos. Class Acc: action classification accuracy. Avg tIoU: average tIoU. Pos Acc: accuracy of positive predictions.}
\label{fig:parameter_curve}
\end{figure}

We also investigate the hyperparameters in ICD. Fig. \ref{fig:parameter_curve}(a) shows the performance curve of average mAP corresponding to threshold $\tau_{icd}$ on THUMOS14 with 10\% labeled data. The average mAP gradually improves as $\tau_{icd}$ increases, but it slightly drops when $\tau_{icd}$ reaches 0.3. This is because setting $\tau_{icd}$ too large may lead to the removal of more true positives. Fig. \ref{fig:parameter_curve}(b) presents the average mAP for different values of $\varsigma_{icd}$. Our method achieves the highest performance when $\varsigma_{icd}$ is set to 0.7.
Thus, we set $\tau_{icd}$ to 0.3 and $\varsigma_{icd}$ to 0.7.

\noindent\textbf{Ablation on ACP.} To go deeper into our ACP, we conducted three experiments:
coarse-grained contrast loss only, fine-grained contras loss only, and the complete pre-training loss. From Tab. \ref{tab:ACP_loss}, either $\mathcal{L}_{conc}$ or $\mathcal{L}_{conf}$ can improve the performance compared to the baseline (No ACP), and when we combine 
them together, the average mAP experiences a notable increase of 1.2\% and 0.9\% in the case of the 10\% and 40\% settings, respectively. This improvement can be attributed to the enhanced capability of the model to discriminate between actions and backgrounds, as well as between different actions.

\begin{table}[tp]  
\begin{minipage}{0.5\linewidth}
\captionof{table}{The effect of ICD on THUMOS14 with 10\% and 40\% labeled videos. EAP and MPP refer to eliminating ambiguous positives and mining potential positives.}
\centering          
\resizebox{0.9\textwidth}{1.8cm}{\begin{tabular}{c|c|ccc|c}
\toprule
\multirow{2}{*}{Label} & \multirow{2}{*}{Method} &  \multicolumn{4}{c}{mAP(\%)} 
\\ \cmidrule{3-6}
 &  & 0.3 & 0.5 & 0.7 & Avg. \\ \midrule
\multirow{4}{*}{ 10\%} & No ICD  &31.9&22.9 & 8.1 &21.3\\
& EAP & 31.7 & 23.6 & 9.9 & 22.1 \\
&MPP &32.4 & 23.5 & 10.0 & 22.5\\
&EAP+MPP &\textbf{33.6} & \textbf{24.6} & \textbf{10.2} & \textbf{23.3}\\
\midrule
\multirow{4}{*}{ 40\%}  & No ICD &71.0 &56.7 &26.8 & 52.6 \\
&EAP & 71.5 & 57.3 &27.2 & 53.2 \\
&MPP & 71.7 & 57.7 & 27.5 & 53.3\\
&EAP+MPP  &\textbf{72.0} & \textbf{58.1} & \textbf{27.7} & \textbf{53.9}\\
\bottomrule
\end{tabular}}
\label{tab:ICD}
\end{minipage}
\hfill
\begin{minipage}{0.48\linewidth}
\caption{The effect of coarse- and fine-grained contrasts on THUMOS14 with 10\% and 40\%  labeled videos.}
\centering          
\resizebox{0.9\textwidth}{1.8cm}{\begin{tabular}{c|c|ccc|c}
\toprule
\multirow{2}{*}{Label} & \multirow{2}{*}{Loss} &  \multicolumn{4}{c}{mAP(\%)} 
\\ \cmidrule{3-6}
 &  & 0.3 & 0.5 & 0.7 & Avg. \\ \midrule
\multirow{4}{*}{ 10\%} & No ACP  &33.6 & 24.6 & 10.2 & 23.3\\
&$\mathcal{L}_{conc}$ &34.6 &25.5 &10.6 & 24.1 \\
&$\mathcal{L}_{conf}$ &34.4 & 25.1 & 10.5 & 23.8\\
&$\mathcal{L}_{conc}$+$\mathcal{L}_{conf}$ & \textbf{35.1} & \textbf{25.6} & \textbf{11.0}& \textbf{24.5}\\
\midrule
\multirow{4}{*}{ 40\%} & No ACP &72.0 & 58.1 & 27.7 & 53.9\\
&$\mathcal{L}_{conc}$ & 72.5 & 58.6 & 28.4 & 54.4\\
&$\mathcal{L}_{conf}$ & 72.3 & 59.0 & 27.9 &54.4\\
&$\mathcal{L}_{conc}$+$\mathcal{L}_{conf}$  &\textbf{73.2} &  \textbf{59.1} & \textbf{28.7} & \textbf{54.8}\\
\bottomrule
\end{tabular}}
\label{tab:ACP_loss}
\end{minipage}
\end{table}

\noindent\textbf{Ablation on the quality of pseudo labels.} We further study the quality of pseudo labels in terms of classification accuracy (Class Acc), average temporal IoU (Avg tIoU) w.r.t ground truth and accuracy of positive predictions (Pos Acc). Here positive predictions mean that the estimated instance has the same action class as the ground truth and tIoU is above 0.5. 
Specifically, we consider three cases: first, where $\tau_{pos}$ in Eq. (\ref{equation:split_instance}) is fixed at 0.3 to filter pseudo labels; second, where $\tau_{pos}$ is dynamically computed; third, 
where the pseudo labels are enhanced by EAP, MPP and EAP+MPP, respectively. Note that the results reported in Fig. \ref{fig:parameter_curve}(c) are calculated based on 90\% unlabeled videos. From Fig. \ref{fig:parameter_curve}(c), we can observe that the quality of pseudo labels is very poor across all metrics when using a fixed threshold, while it is improved by a large margin after employing a dynamical threshold.  Based on the positives and negatives initially divided by the dynamical $\tau_{pos}$, both EAP and MPP contribute to improving the pseudo-label quality and their combination leads to more reliable pseudo labels. 
\begin{figure*}[tp]
\begin{center}
\includegraphics[width=10cm]{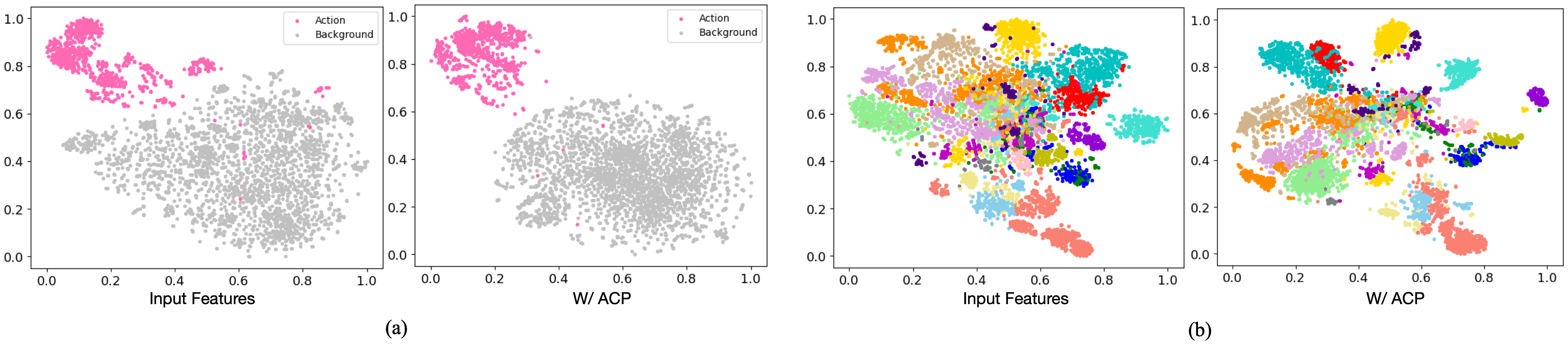}
\end{center}
\caption{The effect of our ACP on THUMOS14. (a) t-SNE visualization of action and background features. (b)  t-SNE visualization of features for different actions. The legend for different actions is provided in the supplementary material.}
\label{fig:tsne}
\end{figure*}

\noindent\textbf{Qualitative results.} To better illustrate the effectiveness of ACP, we visualize some qualitative results on THUMOS14 in Fig. \ref{fig:tsne}. Specifically, we choose a subset of the validation set and visualize the learned representation (Sec. \ref{sec:3.3}). In Fig.  \ref{fig:tsne}(a), the visualization demonstrates better separation between action and background features, indicating that ACP effectively enhances the discrimination between action-related and background information. Additionally, Fig. \ref{fig:tsne}(b) shows the result where distinct clusters can be observed for each action category. This suggests that ACP contributes to improved feature representation, allowing for better discrimination between different actions. Please refer to \textbf{Supp. E} for more ablation studies and visualization results.

\section{Conclusion}
In this paper, we explore the label noise problem in SS-TAL with a novel framework. We introduce ALQA to jointly learn classification confidence and localization reliability, providing more reliable joint scores for pseudo-label ranking. Additionally, ICD is designed to improve pseudo-label quality by eliminating ambiguous positives and mining potential positives. Finally, we propose ACP pre-training to enhance discrimination within and between actions and backgrounds, serving as a versatile component for SS-TAL methods. Extensive experiments on two benchmarks demonstrate new state-of-the-art performance.

\textbf{Limitation and future work.} In the ICD, different similarity score thresholds may affect the quality of pseudo-label refinement. Exploring more adaptive strategies to identify positive instances is a potential avenue for future work. Additionally, the generalization ability of our approach across different TAL frameworks could be further explored.

\section*{Acknowledgements}
The research is supported by TAILOR, a project funded by EU Horizon 2020 research and innovation programme under GA No 952215.

%
%
\bibliographystyle{splncs04}
\bibliography{main}

\begin{thebibliography}{10}
\providecommand{\url}[1]{\texttt{#1}}
\providecommand{\urlprefix}{URL }
\providecommand{\doi}[1]{https://doi.org/#1}

\bibitem{bai2020boundary}
Bai, Y., Wang, Y., Tong, Y., Yang, Y., Liu, Q., Liu, J.: Boundary content graph neural network for temporal action proposal generation. In: Computer Vision--ECCV 2020: 16th European Conference, Glasgow, UK, August 23--28, 2020, Proceedings, Part XXVIII 16. pp. 121--137. Springer (2020)

\bibitem{berthelot2019mixmatch}
Berthelot, D., Carlini, N., Goodfellow, I., Papernot, N., Oliver, A., Raffel, C.A.: Mixmatch: A holistic approach to semi-supervised learning. Advances in neural information processing systems  \textbf{32} (2019)

\bibitem{bodla2017soft}
Bodla, N., Singh, B., Chellappa, R., Davis, L.S.: Soft-nms--improving object detection with one line of code. In: Proceedings of the IEEE international conference on computer vision. pp. 5561--5569 (2017)

\bibitem{caba2015activitynet}
Caba~Heilbron, F., Escorcia, V., Ghanem, B., Carlos~Niebles, J.: Activitynet: A large-scale video benchmark for human activity understanding. In: Proceedings of the ieee conference on computer vision and pattern recognition. pp. 961--970 (2015)

\bibitem{cao2023contrastive}
Cao, S., Joshi, D., Gui, L.Y., Wang, Y.X.: Contrastive mean teacher for domain adaptive object detectors. In: Proceedings of the IEEE/CVF Conference on Computer Vision and Pattern Recognition. pp. 23839--23848 (2023)

\bibitem{carion2020end}
Carion, N., Massa, F., Synnaeve, G., Usunier, N., Kirillov, A., Zagoruyko, S.: End-to-end object detection with transformers. In: European conference on computer vision. pp. 213--229. Springer (2020)

\bibitem{carreira2017quo}
Carreira, J., Zisserman, A.: Quo vadis, action recognition? a new model and the kinetics dataset. In: proceedings of the IEEE Conference on Computer Vision and Pattern Recognition. pp. 6299--6308 (2017)

\bibitem{chen2022label}
Chen, B., Chen, W., Yang, S., Xuan, Y., Song, J., Xie, D., Pu, S., Song, M., Zhuang, Y.: Label matching semi-supervised object detection. In: Proceedings of the IEEE/CVF Conference on Computer Vision and Pattern Recognition. pp. 14381--14390 (2022)

\bibitem{chen2022dense}
Chen, B., Li, P., Chen, X., Wang, B., Zhang, L., Hua, X.S.: Dense learning based semi-supervised object detection. In: Proceedings of the IEEE/CVF conference on computer vision and pattern recognition. pp. 4815--4824 (2022)

\bibitem{chen2020simple}
Chen, T., Kornblith, S., Norouzi, M., Hinton, G.: A simple framework for contrastive learning of visual representations. In: International conference on machine learning. pp. 1597--1607. PMLR (2020)

\bibitem{damen2022rescaling}
Damen, D., Doughty, H., Farinella, G.M., Furnari, A., Kazakos, E., Ma, J., Moltisanti, D., Munro, J., Perrett, T., Price, W., et~al.: Rescaling egocentric vision: Collection, pipeline and challenges for epic-kitchens-100. International Journal of Computer Vision pp. 1--23 (2022)

\bibitem{escorcia2016daps}
Escorcia, V., Caba~Heilbron, F., Niebles, J.C., Ghanem, B.: Daps: Deep action proposals for action understanding. In: Computer Vision--ECCV 2016: 14th European Conference, Amsterdam, The Netherlands, October 11-14, 2016, Proceedings, Part III 14. pp. 768--784. Springer (2016)

\bibitem{gao2018ctap}
Gao, J., Chen, K., Nevatia, R.: Ctap: Complementary temporal action proposal generation. In: Proceedings of the European conference on computer vision (ECCV). pp. 68--83 (2018)

\bibitem{gutmann2010noise}
Gutmann, M., Hyv{\"a}rinen, A.: Noise-contrastive estimation: A new estimation principle for unnormalized statistical models. In: Proceedings of the thirteenth international conference on artificial intelligence and statistics. pp. 297--304. JMLR Workshop and Conference Proceedings (2010)

\bibitem{he2020momentum}
He, K., Fan, H., Wu, Y., Xie, S., Girshick, R.: Momentum contrast for unsupervised visual representation learning. In: Proceedings of the IEEE/CVF conference on computer vision and pattern recognition. pp. 9729--9738 (2020)

\bibitem{henaff2021efficient}
H{\'e}naff, O.J., Koppula, S., Alayrac, J.B., Van~den Oord, A., Vinyals, O., Carreira, J.: Efficient visual pretraining with contrastive detection. In: Proceedings of the IEEE/CVF International Conference on Computer Vision. pp. 10086--10096 (2021)

\bibitem{hyvarinen2005estimation}
Hyv{\"a}rinen, A., Dayan, P.: Estimation of non-normalized statistical models by score matching. Journal of Machine Learning Research  \textbf{6}(4) (2005)

\bibitem{ji2019learning}
Ji, J., Cao, K., Niebles, J.C.: Learning temporal action proposals with fewer labels. In: Proceedings of the IEEE/CVF International Conference on Computer Vision. pp. 7073--7082 (2019)

\bibitem{THUMOS14}
Jiang, Y.G., Liu, J., Roshan~Zamir, A., Toderici, G., Laptev, I., Shah, M., Sukthankar, R.: {THUMOS} challenge: Action recognition with a large number of classes. \url{http://crcv.ucf.edu/THUMOS14/} (2014)

\bibitem{laine2016temporal}
Laine, S., Aila, T.: Temporal ensembling for semi-supervised learning. arXiv preprint arXiv:1610.02242  (2016)

\bibitem{lee2013pseudo}
Lee, D.H., et~al.: Pseudo-label: The simple and efficient semi-supervised learning method for deep neural networks. In: Workshop on challenges in representation learning, ICML. vol.~3, p.~896. Atlanta (2013)

\bibitem{li2023semi}
Li, P., Purkait, P., Ajanthan, T., Abdolshah, M., Garg, R., Husain, H., Xu, C., Gould, S., Ouyang, W., van~den Hengel, A.: Semi-supervised semantic segmentation under label noise via diverse learning groups. In: Proceedings of the IEEE/CVF International Conference on Computer Vision. pp. 1229--1238 (2023)

\bibitem{li2017joint}
Li, W., Wang, W., Chen, X., Wang, J., Li, G.: A joint model for action localization and classification in untrimmed video with visual attention. In: 2017 IEEE International Conference on Multimedia and Expo (ICME). pp. 619--624. IEEE (2017)

\bibitem{li2020generalized}
Li, X., Wang, W., Wu, L., Chen, S., Hu, X., Li, J., Tang, J., Yang, J.: Generalized focal loss: Learning qualified and distributed bounding boxes for dense object detection. Advances in Neural Information Processing Systems  \textbf{33},  21002--21012 (2020)

\bibitem{lin2020fast}
Lin, C., Li, J., Wang, Y., Tai, Y., Luo, D., Cui, Z., Wang, C., Li, J., Huang, F., Ji, R.: Fast learning of temporal action proposal via dense boundary generator. In: Proceedings of the AAAI conference on artificial intelligence. vol.~34, pp. 11499--11506 (2020)

\bibitem{lin2021learning}
Lin, C., Xu, C., Luo, D., Wang, Y., Tai, Y., Wang, C., Li, J., Huang, F., Fu, Y.: Learning salient boundary feature for anchor-free temporal action localization. In: Proceedings of the IEEE/CVF Conference on Computer Vision and Pattern Recognition. pp. 3320--3329 (2021)

\bibitem{lin2019bmn}
Lin, T., Liu, X., Li, X., Ding, E., Wen, S.: Bmn: Boundary-matching network for temporal action proposal generation. In: Proceedings of the IEEE/CVF international conference on computer vision. pp. 3889--3898 (2019)

\bibitem{lin2017single}
Lin, T., Zhao, X., Shou, Z.: Single shot temporal action detection. In: Proceedings of the 25th ACM international conference on Multimedia. pp. 988--996 (2017)

\bibitem{lin2018bsn}
Lin, T., Zhao, X., Su, H., Wang, C., Yang, M.: Bsn: Boundary sensitive network for temporal action proposal generation. In: Proceedings of the European conference on computer vision (ECCV). pp. 3--19 (2018)

\bibitem{liu2023ambiguity}
Liu, C., Zhang, W., Lin, X., Zhang, W., Tan, X., Han, J., Li, X., Ding, E., Wang, J.: Ambiguity-resistant semi-supervised learning for dense object detection. In: Proceedings of the IEEE/CVF Conference on Computer Vision and Pattern Recognition. pp. 15579--15588 (2023)

\bibitem{liu2023mixteacher}
Liu, L., Zhang, B., Zhang, J., Zhang, W., Gan, Z., Tian, G., Zhu, W., Wang, Y., Wang, C.: Mixteacher: Mining promising labels with mixed scale teacher for semi-supervised object detection. In: Proceedings of the IEEE/CVF Conference on Computer Vision and Pattern Recognition. pp. 7370--7379 (2023)

\bibitem{miyato2018virtual}
Miyato, T., Maeda, S.i., Koyama, M., Ishii, S.: Virtual adversarial training: a regularization method for supervised and semi-supervised learning. IEEE transactions on pattern analysis and machine intelligence  \textbf{41}(8),  1979--1993 (2018)

\bibitem{nag2022semi}
Nag, S., Zhu, X., Song, Y.Z., Xiang, T.: Semi-supervised temporal action detection with proposal-free masking. In: European Conference on Computer Vision. pp. 663--680. Springer (2022)

\bibitem{oord2018representation}
Oord, A.v.d., Li, Y., Vinyals, O.: Representation learning with contrastive predictive coding. arXiv preprint arXiv:1807.03748  (2018)

\bibitem{ouali2020semi}
Ouali, Y., Hudelot, C., Tami, M.: Semi-supervised semantic segmentation with cross-consistency training. In: Proceedings of the IEEE/CVF Conference on Computer Vision and Pattern Recognition. pp. 12674--12684 (2020)

\bibitem{qing2021temporal}
Qing, Z., Su, H., Gan, W., Wang, D., Wu, W., Wang, X., Qiao, Y., Yan, J., Gao, C., Sang, N.: Temporal context aggregation network for temporal action proposal refinement. In: Proceedings of the IEEE/CVF conference on computer vision and pattern recognition. pp. 485--494 (2021)

\bibitem{redmon2016you}
Redmon, J., Divvala, S., Girshick, R., Farhadi, A.: You only look once: Unified, real-time object detection. In: Proceedings of the IEEE conference on computer vision and pattern recognition. pp. 779--788 (2016)

\bibitem{ren2015faster}
Ren, S., He, K., Girshick, R., Sun, J.: Faster r-cnn: Towards real-time object detection with region proposal networks. Advances in neural information processing systems  \textbf{28} (2015)

\bibitem{shao2023action}
Shao, J., Wang, X., Quan, R., Zheng, J., Yang, J., Yang, Y.: Action sensitivity learning for temporal action localization. arXiv preprint arXiv:2305.15701  (2023)

\bibitem{shi2021temporal}
Shi, B., Dai, Q., Hoffman, J., Saenko, K., Darrell, T., Xu, H.: Temporal action detection with multi-level supervision. In: Proceedings of the IEEE/CVF International Conference on Computer Vision. pp. 8022--8032 (2021)

\bibitem{shi2023tridet}
Shi, D., Zhong, Y., Cao, Q., Ma, L., Li, J., Tao, D.: Tridet: Temporal action detection with relative boundary modeling. In: Proceedings of the IEEE/CVF Conference on Computer Vision and Pattern Recognition. pp. 18857--18866 (2023)

\bibitem{shi2022react}
Shi, D., Zhong, Y., Cao, Q., Zhang, J., Ma, L., Li, J., Tao, D.: React: Temporal action detection with relational queries. In: European conference on computer vision. pp. 105--121. Springer (2022)

\bibitem{singh2021semi}
Singh, A., Chakraborty, O., Varshney, A., Panda, R., Feris, R., Saenko, K., Das, A.: Semi-supervised action recognition with temporal contrastive learning. In: Proceedings of the IEEE/CVF Conference on Computer Vision and Pattern Recognition. pp. 10389--10399 (2021)

\bibitem{sohn2020fixmatch}
Sohn, K., Berthelot, D., Carlini, N., Zhang, Z., Zhang, H., Raffel, C.A., Cubuk, E.D., Kurakin, A., Li, C.L.: Fixmatch: Simplifying semi-supervised learning with consistency and confidence. Advances in neural information processing systems  \textbf{33},  596--608 (2020)

\bibitem{su2023multi}
Su, T., Wang, H., Wang, L.: Multi-level content-aware boundary detection for temporal action proposal generation. IEEE Transactions on Image Processing  (2023)

\bibitem{tang2021humble}
Tang, Y., Chen, W., Luo, Y., Zhang, Y.: Humble teachers teach better students for semi-supervised object detection. In: Proceedings of the IEEE/CVF Conference on Computer Vision and Pattern Recognition. pp. 3132--3141 (2021)

\bibitem{tarvainen2017mean}
Tarvainen, A., Valpola, H.: Mean teachers are better role models: Weight-averaged consistency targets improve semi-supervised deep learning results. Advances in neural information processing systems  \textbf{30} (2017)

\bibitem{tian2019fcos}
Tian, Z., Shen, C., Chen, H., He, T.: Fcos: Fully convolutional one-stage object detection. In: Proceedings of the IEEE/CVF international conference on computer vision. pp. 9627--9636 (2019)

\bibitem{wang2016temporal}
Wang, L., Xiong, Y., Wang, Z., Qiao, Y., Lin, D., Tang, X., Van~Gool, L.: Temporal segment networks: Towards good practices for deep action recognition. In: European conference on computer vision. pp. 20--36. Springer (2016)

\bibitem{wang2021self}
Wang, X., Zhang, S., Qing, Z., Shao, Y., Gao, C., Sang, N.: Self-supervised learning for semi-supervised temporal action proposal. In: Proceedings of the IEEE/CVF Conference on Computer Vision and Pattern Recognition. pp. 1905--1914 (2021)

\bibitem{wang2018iterative}
Wang, Y., Liu, W., Ma, X., Bailey, J., Zha, H., Song, L., Xia, S.T.: Iterative learning with open-set noisy labels. In: Proceedings of the IEEE conference on computer vision and pattern recognition. pp. 8688--8696 (2018)

\bibitem{weng2022efficient}
Weng, Y., Pan, Z., Han, M., Chang, X., Zhuang, B.: An efficient spatio-temporal pyramid transformer for action detection. In: European Conference on Computer Vision. pp. 358--375. Springer (2022)

\bibitem{xia2023learning}
Xia, K., Wang, L., Zhou, S., Hua, G., Tang, W.: Learning from noisy pseudo labels for semi-supervised temporal action localization. In: Proceedings of the IEEE/CVF International Conference on Computer Vision. pp. 10160--10169 (2023)

\bibitem{xie2021detco}
Xie, E., Ding, J., Wang, W., Zhan, X., Xu, H., Sun, P., Li, Z., Luo, P.: Detco: Unsupervised contrastive learning for object detection. In: Proceedings of the IEEE/CVF International Conference on Computer Vision. pp. 8392--8401 (2021)

\bibitem{xie2023boosting}
Xie, H., Wang, C., Zheng, M., Dong, M., You, S., Fu, C., Xu, C.: Boosting semi-supervised semantic segmentation with probabilistic representations. In: Proceedings of the AAAI Conference on Artificial Intelligence. vol.~37, pp. 2938--2946 (2023)

\bibitem{xing2023svformer}
Xing, Z., Dai, Q., Hu, H., Chen, J., Wu, Z., Jiang, Y.G.: Svformer: Semi-supervised video transformer for action recognition. In: Proceedings of the IEEE/CVF Conference on Computer Vision and Pattern Recognition. pp. 18816--18826 (2023)

\bibitem{xu2022contrastive}
Xu, M., Gundogdu, E., Lapin, M., Ghanem, B., Donoser, M., Bazzani, L.: Contrastive language-action pre-training for temporal localization. arXiv preprint arXiv:2204.12293  (2022)

\bibitem{yang2022class}
Yang, F., Wu, K., Zhang, S., Jiang, G., Liu, Y., Zheng, F., Zhang, W., Wang, C., Zeng, L.: Class-aware contrastive semi-supervised learning. In: Proceedings of the IEEE/CVF Conference on Computer Vision and Pattern Recognition. pp. 14421--14430 (2022)

\bibitem{yang2022temporal}
Yang, H., Wu, W., Wang, L., Jin, S., Xia, B., Yao, H., Huang, H.: Temporal action proposal generation with background constraint. In: Proceedings of the AAAI conference on artificial intelligence. vol.~36, pp. 3054--3062 (2022)

\bibitem{yang2023revisiting}
Yang, L., Qi, L., Feng, L., Zhang, W., Shi, Y.: Revisiting weak-to-strong consistency in semi-supervised semantic segmentation. In: Proceedings of the IEEE/CVF Conference on Computer Vision and Pattern Recognition. pp. 7236--7246 (2023)

\bibitem{zeng2019graph}
Zeng, R., Huang, W., Tan, M., Rong, Y., Zhao, P., Huang, J., Gan, C.: Graph convolutional networks for temporal action localization. In: Proceedings of the IEEE/CVF international conference on computer vision. pp. 7094--7103 (2019)

\bibitem{zhang2022actionformer}
Zhang, C.L., Wu, J., Li, Y.: Actionformer: Localizing moments of actions with transformers. In: European Conference on Computer Vision. pp. 492--510. Springer (2022)

\bibitem{zhang2017mixup}
Zhang, H., Cisse, M., Dauphin, Y.N., Lopez-Paz, D.: mixup: Beyond empirical risk minimization. arXiv preprint arXiv:1710.09412  (2017)

\bibitem{zhao2021video}
Zhao, C., Thabet, A.K., Ghanem, B.: Video self-stitching graph network for temporal action localization. In: Proceedings of the IEEE/CVF International Conference on Computer Vision. pp. 13658--13667 (2021)

\bibitem{zhao2020bottom}
Zhao, P., Xie, L., Ju, C., Zhang, Y., Wang, Y., Tian, Q.: Bottom-up temporal action localization with mutual regularization. In: Computer Vision--ECCV 2020: 16th European Conference, Glasgow, UK, August 23--28, 2020, Proceedings, Part VIII 16. pp. 539--555. Springer (2020)

\bibitem{zheng2020distance}
Zheng, Z., Wang, P., Liu, W., Li, J., Ye, R., Ren, D.: Distance-iou loss: Faster and better learning for bounding box regression. In: Proceedings of the AAAI conference on artificial intelligence. vol.~34, pp. 12993--13000 (2020)

\bibitem{zhou2023smc}
Zhou, F., Jiang, Z., Zhou, H., Li, X.: Smc-nca: Semantic-guided multi-level contrast for semi-supervised action segmentation. arXiv preprint arXiv:2312.12347  (2023)

\bibitem{zhu2021enriching}
Zhu, Z., Tang, W., Wang, L., Zheng, N., Hua, G.: Enriching local and global contexts for temporal action localization. In: Proceedings of the IEEE/CVF international conference on computer vision. pp. 13516--13525 (2021)

\end{thebibliography}

\clearpage
\onecolumn

\setcounter{table}{0}
\setcounter{figure}{0}
\setcounter{equation}{0}
\setcounter{page}{1}
\renewcommand\thefigure{S\arabic{figure}}
\renewcommand\thetable{S\arabic{table}}

\section*{Supplementary Material: Towards Adaptive Pseudo-label Learning for Semi-Supervised Temporal Action Localization}

\subsection*{A. More implementation details. }
For THUMOS14, the initial learning rate is set to 1e-4, and a cosine learning rate decay is used. The mini-batch size and weight decay are set to 2 and 1e-4, respectively. For ActivityNet v1.3, the learning rate, mini-batch size and weight decay are set to 1e-3, 16 and 1e-4, respectively.

For semi-supervised learning, we first pre-train the baseline on the labeled and unlabeled videos for 15 epochs without using any GT labels. Secondly, based on the pre-trained model, we update it on the labeled videos by using GT labels and then apply it to unlabeled videos to obtain pseudo labels.  
Finally, the model can be further updated by jointly optimizing supervised loss and unsupervised loss for 40 epochs on THUMOS14 and 15 epochs on ActivityNet v1.3.  Notably, we train the baseline and ICD together in the second stage. The number of instances for training ICD is set to 10, and $\tau_{icd}$ and $\varsigma_{icd}$ are set to 0.3 and 0.7, respectively. Besides, the number of clusters $B$ of fine-grained contrast is set to 4 and 2 for THUMOS14 and ActivityNet v1.3, respectively. All source code will be made publicly available.

\subsection*{B. More Related Works. }

\noindent\textbf{Contrastive Learning.} Contrastive learning \cite{oord2018representation,chen2020simple,he2020momentum} has been widely explored for self-supervised visual representation learning with promising results. The main objective of contrastive learning is to learn effective representations by encouraging the network to distinguish between different instances in feature space while pulling the same instance closer. \cite{hyvarinen2005estimation,gutmann2010noise} established a foundation for contrast learning by proposing and improving Noise Contrast Estimation (NCE). Based on NCE, infoNCE \cite{oord2018representation} aims to learn representations by maximizing the mutual information between positive samples (perturbed views of the same input) while minimizing the mutual information between positive and negative samples (perturbed views of different inputs), which has been successfully applied to different tasks, such as object detection\cite{xie2021detco,cao2023contrastive}, semantic segmentation \cite{henaff2021efficient,xie2023boosting}  and temporal action localization \cite{lin2021learning,nag2022semi,shi2022react}. 
In TAL, \cite{lin2021learning} leverages contrastive learning to help with the learning of boundary features while \cite{shi2022react} contrasts different actions based on the aggregated representation of action segments. Unlike these methods designed for supervised learning, we develop a unsupervised contrastive pre-training scheme for SS-TAL to enhance frame-level representation, which benefits the subsequent semi-supervised learning using only a small amount of labeled videos.  \cite{xu2022contrastive} employs vision-language contrastive learning, while our Action-aware Contrastive Pretraining (ACP) operates independently of language descriptions. Unlike \cite{xu2022contrastive}, which leverages language data to capture visio-linguistic relations, our ACP focuses on enhancing frame-level representation by performing coarse- and fine-grained contrasts to improve discrimination both within actions and between actions and backgrounds. Additionally, ACP can be seamlessly integrated with various backbones (Tab.\ref{tab:ablation_ACP_spot}).

\subsection*{C. Framework Diagram for Action-aware Contrastive Pre-training}
\begin{figure*}[htb]
\begin{center}
\includegraphics[width=10.5cm]{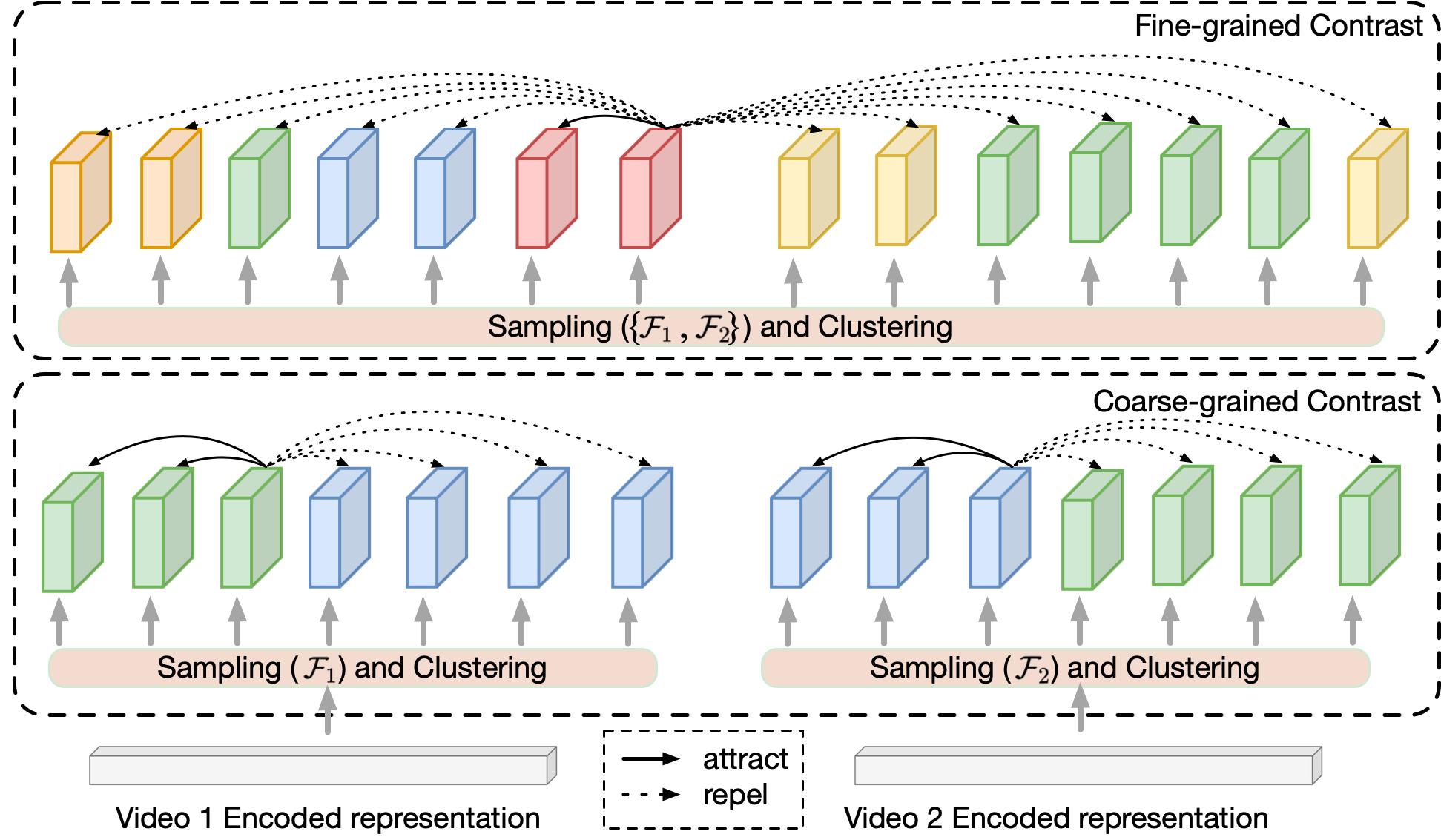}
\end{center}
\caption{Illustration of our Action-aware Contrastive Pre-training. We first obtain temporal representation encoded by the base model (\eg, ActionFormer \cite{zhang2022actionformer}). In coarse-grained contrast, two videos within a mini-batch are sampled to form representation set $\mathcal{F}_{1}$ and $\mathcal{F}_{2}$, respectively and we cluster the corresponding input features to generate frame-wise clustering labels with only 2 action categories (0-action,1-background). We then contrast between actions and backgrounds to attract similar representations and repel different representations. In fine-grained contrast, we contrast more between different kinds of actions based on the combined representation set $\{\mathcal{F}_{1},\mathcal{F}_{2}\}$.}
\label{fig:contrastive}
\end{figure*}

To help understand our Action-aware Contrastive Pre-training (ACP) better, we provide a detailed framework diagram, as shown in Fig. \ref{fig:contrastive}.

\subsection*{D. The legend for different actions}

\begin{figure}[H]
\begin{center}
\includegraphics[width=12.5cm]{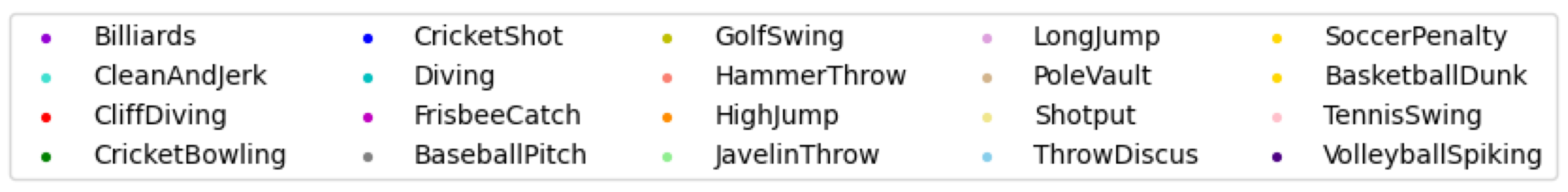}
\end{center}
\caption{The legend for different actions in Fig. 4(b).}
\label{fig:tsne_legend}
\end{figure}

We provide the legend for different actions, as shown in Fig. \ref{fig:tsne_legend}.

\subsection*{E. More Ablation Studies and Comparisons}
\begin{table}[ht]
\caption{The effect of different loss functions in  Eq. (3).}
\centering
    \setlength{\tabcolsep}{3mm}
    \begin{tabular}{lccccc}
        \hline
        Label &Method & 0.3 & 0.5 & 0.7 & Avg \\
        \hline
        \multirow{2}{*}{ 10\%}
        &ALQA w/ KL divergence & 31.2 & 21.9 & \textbf{8.1} & 20.8 \\ 
        & ALQA w/ BCE &\textbf{31.9} &\textbf{22.9} &\textbf{8.1}  & \textbf{21.3} \\ 
        \midrule
        \multirow{2}{*}{ 40\%}
        &ALQA w/ KL divergence & 70.8 & 56.3 & 26.6 & 52.3  \\ 
        & ALQA w/ BCE  &\textbf{71.0} &\textbf{56.7} &\textbf{26.8} &\textbf{52.6} \\ 
        \hline
    \end{tabular}
    \label{fig:KL_loss}
\end{table}
\textbf{Ablation on BCE and KL divergence loss.} We compare the performance of our method using BCE loss against KL divergence loss (in Eq. (3), as presented in Tab. \ref{fig:KL_loss}, demonstrating that BCE loss achieves a slight improvement in performance over KL divergence loss.

\textbf{Ablation on ACP.} In addition to t-SNE visualization in Fig. 4, we report the accuracy of a linear classifier on the learned features/representations and original input features ( \ie, I3D features extracted by pre-trained model), as shown in Tab. \ref{tab:liner}. 

\begin{table}[h]
\caption{The effect of ACP.}
\centering
    \begin{tabular}{lcc}
        \hline
        Method & Accuracy ($\%$) & Recall ($\%$)\\
        \hline
        Input features & 72.1 & 66.3\\ 
        Learned features with ACP & 76.2 & 73.0 \\ 
        \hline
    \end{tabular}
    \label{tab:liner}
\end{table}

\textbf{Analysis on the generalizability of ACP.}
To validate the generalizability of our proposed ACP, we also add it to the BMN \cite{lin2019bmn} and SPOT \cite{zhang2022actionformer} frameworks. The experimental results are shown in Tab. \ref{tab:ablation_ACP_spot}, which reveal that ACP can significantly improve the performance of the two frameworks. 
Additionally, when comparing our ACP with PT \cite{zhang2022actionformer}, our method outperforms PT by 1\% in the average mAP for BMN. The result remains comparable even when we replace the PT with our ACP in SPOT. However, it's worth noting that our method's performance is slightly lower compared to the PT-based method. This discrepancy can be attributed to the fact that SPOT contains a mask head to predict action boundaries, while PT also incorporates mask learning loss. Therefore, PT seems to be more specifically tailored for SPOT. In contrast, our ACP is designed to enhance frame-level representation and does not rely on the model's architecture.

\begin{table}[ht]
\caption{The generalizability of ACP under different frameworks (using TSN) on ActivityNet V1.3 with 10\% labeled videos. PT: pre-training method in SPOT \cite{nag2022semi}.}
\centering            
\begin{tabular}{c|ccc|c}
\toprule
 \multirow{2}{*}{Framework} &  \multicolumn{4}{c}{mAP(\%)} 
\\ \cmidrule{2-5}
& 0.5 & 0.75 & 0.95 & Avg. \\ \midrule
BMN \cite{lin2019bmn} (w/o PT) &35.4 &26.4 &8.0 & 25.8\\
BMN \cite{lin2019bmn} (w/ PT)&36.2 &- &- &26.3\\
BMN \cite{lin2019bmn} (w/ ACP)& \textbf{37.6} &27.9 &8.8 &\textbf{27.3}\\
\midrule
SPOT \cite{nag2022semi} (w/o PT) &46.2 &-&-&30.5\\
SPOT \cite{nag2022semi} (w/ PT)&\textbf{49.9} &\textbf{31.1}&\textbf{8.3}&\textbf{32.1}\\
SPOT \cite{nag2022semi} (w/ ACP)  & 48.8 & 30.9 & 8.2 & 31.9\\
\bottomrule
\end{tabular}
\label{tab:ablation_ACP_spot}
\end{table}

\textbf{Ablation on the pre-trained loss $\mathcal{L}_{acp}$.} As mentioned in Sec. 3.3, $\mathcal{L}_{acp}$ is utilized not only during the pre-training phase but also in subsequent semi-supervised learning stages. To further investigate its effect on semi-supervised learning, we discuss three cases in Tab. \ref{tab:ablation_ACP_label_unlabel}. Firstly, $\mathcal{L}_{acp}$ is solely employed for pre-training (row 1). Secondly, after pre-training, $\mathcal{L}_{acp}$ is involved in semi-supervised learning, where GT action labels of labeled data replace clustering labels to guide fine-grained and coarse-grained contrastive learning (row 2). Thirdly, after pre-training, $\mathcal{L}_{acp}$ simultaneously operates on labeled and unlabeled data for semi-supervised learning, with contrastive learning guided by GT labels and pseudo labels, respectively (row 3). From Tab. \ref{tab:ablation_ACP_label_unlabel},  we can see an increase in the average mAP from 23.6\% to 24.1\% to 24.5\% when $\mathcal{L}_{acp}$ is utilized in conjunction with labeled data, and then with both labeled and unlabeled data, respectively. This suggests that incorporating $\mathcal{L}_{acp}$ into semi-supervised learning, especially with the inclusion of both labeled and unlabeled data, leads to improved semi-supervised performance.

\begin{table}[h]
\centering
\caption{The effect of pre-training loss (\ie, $\mathcal{L}_{acp}$) during fine-tuning on THUMOS14 with 10\% labeled videos.}
\label{tab:ablation_ACP_label_unlabel}
\begin{tabular}{c|ccc|c}
\toprule
 \multirow{2}{*}{Method} &  \multicolumn{4}{c}{mAP(\%)} 
\\ \cmidrule{2-5}
& 0.3 & 0.5 & 0.7 & Avg. \\ \midrule
$\mathcal{L}_{acp}$ (only pre-training) &34.3&24.7&10.5 &23.6\\
$\mathcal{L}_{acp}$ (labeled data)  & 34.6 & 25.5 & 10.6 & 24.1\\
$\mathcal{L}_{acp}$ (labeled+unlabeled data)  & \textbf{35.1} & \textbf{25.6} & \textbf{11.0}& \textbf{24.5}\\
\bottomrule
\end{tabular}
\end{table}

\textbf{Ablation on the number of clusters in the fine-grained contrast.} Different from the coarse-grained contrast that only involves two action classes, \ie, action and background, fine-grained contrast with a mini-batch normally involves multiple action classes. Thus, the number of clusters is important to split positive and negative pairs for fine-grained contrast. In Tab. \ref{tab:ablation_ACP_batch}, we report the ablation results for the choice of the number of clusters on THUMOS14 (with an average of 1.12
action classes per video). Specifically, we evaluate the learned representations (Sec. 3.3) by training a simple linear classifier (\ie, LogisticRegression() in Pytorch) to classify frame-level action categories (including background). From Tab. \ref{tab:ablation_ACP_batch}, we observe that increasing the number of clusters generally leads to improvements in both accuracy and recall. For instance, when the batch size $B$ is set to 2, increasing the number of clusters from 2 to 4 results in a notable increase of 5.2\% in accuracy and 2.9\% in recall, respectively. This is because when B is set to a small value, the model struggles to learn the differences between action classes. Similarly, with a batch size of 4, increasing the number of clusters from 4 to 8 leads to an increase in accuracy from 72.4\% to 75.0\%. The batch size is set to 2 in our experiments. Therefore, we set B to 4 accordingly.
\begin{table}[h]
\centering
\caption{The effect of the number of clusters $B$ (within fine-grained contrast) on THUMOS14 with 10\% labeled videos.}
\label{tab:ablation_ACP_batch}
\begin{tabular}{c|c|cc}
\toprule

Batch Size & $B$ &Accuracy (\%) & Recall (\%) \\ 
\midrule
\multirow{3}{*}{2}
&2 &71.0 & 70.1  \\
&3 &72.8 & 73.5\\
&4 &\textbf{76.2} & \textbf{73.0}\\
\midrule
\multirow{3}{*}{ 4}
&4 &  72.4  &72.3\\
&5 &73.9 &74.4\\
&8 &\textbf{75.0} &\textbf{73.8}\\
\bottomrule
\end{tabular}
\end{table}

\textbf{Analysis on computational complexity.} We also compare the computational complexity (\ie, the number of parameters and FLOPs) of our method with the baseline \cite{zhang2022actionformer} on THUMOS14 using an input with the shape $2304\times 2048$, where 2304 and 2048 represents temporal length and feature dimension, respectively. As shown in Tab. \ref{tab:ablation_gflops}, the addition of the tIoU and tND branches results in only a minor increase in parameter count and FLOPs of the head. When we apply the ICD to identifying ambiguous positives and potential positives, the number of parameters and FLOPs only increased by 0.5M and 0.005G, respectively, which is still comparable to the baseline.

\begin{table}
\small
\centering
\caption{The effect of tIoU, tND branches and ICD on THUMOS14 in terms of computational complexity, including parameters and FLOPs. Main refers to all parts of the model except the classification and regression heads.}
\label{tab:ablation_gflops}
\begin{tabular}{c|ccc|ccc}
\toprule
 \multirow{2}{*}{Method} &  \multicolumn{3}{c|}{Params (M)} &  \multicolumn{3}{c}{FLOPs (G)} 
\\ \cmidrule{2-7}
& Main & Head & All & Main & Head & All\\ 
\midrule
ActF \cite{zhang2022actionformer} &26.017 &3.179 & 29.196 & 30.608 & 14.422 & 45.030 \\
ActF (w/ IoU\&TD) &26.017 &3.182 & 29.199 & 30.608 & 14.436 & 45.044 \\
ActF (w/ tIoU\&tND) + ICD &26.017 &3.182 & 29.724 & 30.608 & 14.436 & 45.049 \\
\bottomrule
\end{tabular}
\end{table}

\textbf{Experiments on EPIC-KITCHEN 100.}  THUMOS14 and ActivityNet v1.3 datasets used in our evaluation are widely recognized for TAL. We further opt for a large-scale dataset, \ie, EPIC-KITCHEN 100 \cite{damen2022rescaling}, comprising verb and noun sub-tasks. The results presented in Tab. \ref{tab:epic} are based on 10\% labeled data, demonstrating the effectiveness of our proposed method. 

We have compared our method in Tab.1 of the paper with SOTA \emph{semi-supervised} TAL methods. Following \cite{xia2023learning}, we use the same fully-supervised baselines, \ie, ActF and BMN, for a fair comparison. We further investigate the effectiveness of our method using other baselines such as P-GCN. Our experiments on THUMOS14, as depicted in Tab. \ref{tab:pgcn}, further demonstrate the superiority of the proposed method.

\begin{table}
\caption{Comparisons on EPIC-KITCHEN 100.}
\label{tab:epic}
\centering
    \begin{tabular}{l|ccccccc}
        \hline
        Task &Method & 0.1 & 0.2 & 0.3 & 0.4 & 0.5 &  Avg\\
        \hline
        \multirow{2}{*}{Verb}
        & Baseline (ActF \cite{zhang2022actionformer}) & 14.6 & 14.0 & 13.2 & 11.7 & 9.5 & 12.6\\ 
        & Ours (ActF \cite{zhang2022actionformer}) &\textbf{19.6} &\textbf{18.7} &\textbf{17.5} & \textbf{15.8} & \textbf{12.9} &\textbf{16.9} \\ 
        \midrule
        \multirow{2}{*}{Noun}
        &Baseline (ActF \cite{zhang2022actionformer}) & 10.9 & 10.1 & 9.3& 8.2 & 6.2 & 8.9  \\ 
        & Ours (ActF \cite{zhang2022actionformer})&\textbf{15.1 }&\textbf{14.4} &\textbf{13.4} &\textbf{11.8} & \textbf{9.8} & \textbf{12.9} \\ 
        \hline
    \end{tabular}
\end{table}

\begin{table}
\caption{Comparisons using other baselines.}
\label{tab:pgcn}
\centering
    \begin{tabular}{lccccc}
        \hline
        Label &Method & 0.3 & 0.5 & 0.7 & Avg\\
        \hline
        \multirow{2}{*}{ 10\%}
        &P-GCN \cite{zeng2019graph} & 16.9 & 12.0 & 4.8 &  11.2\\ 
        & Ours & \textbf{26.4}  & \textbf{17.5} & \textbf{5.8} & \textbf{16.6}\\ 
        \midrule
        \multirow{2}{*}{ 40\%}
        &P-GCN  \cite{zeng2019graph} & 52.5 & 38.7 & 14.4 & 35.2 \\ 
        & Ours & \textbf{58.8} & \textbf{43.1} &\textbf{17.3} & \textbf{39.7}\\ 
        \hline
    \end{tabular}
\end{table}

\begin{figure*}
\begin{center}
\includegraphics[width=11cm]{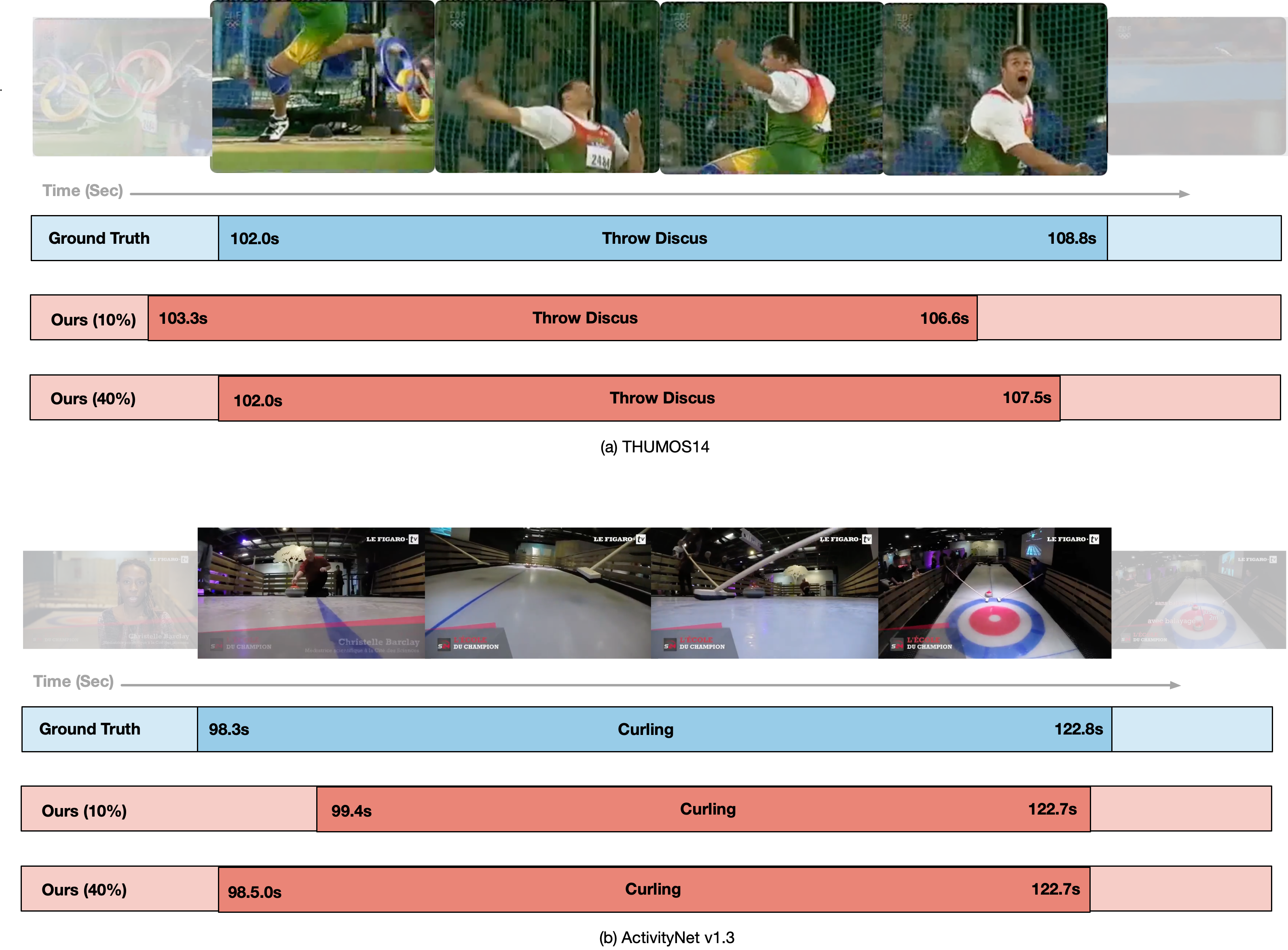}
\end{center}
\caption{Qualitative SS-TAL results on (a) THUMOS14 and (b) ActivityNet v1.3, respectively.}
\label{fig:vis_results}
\end{figure*}

\begin{figure*}
\begin{center}
\includegraphics[width=11cm]{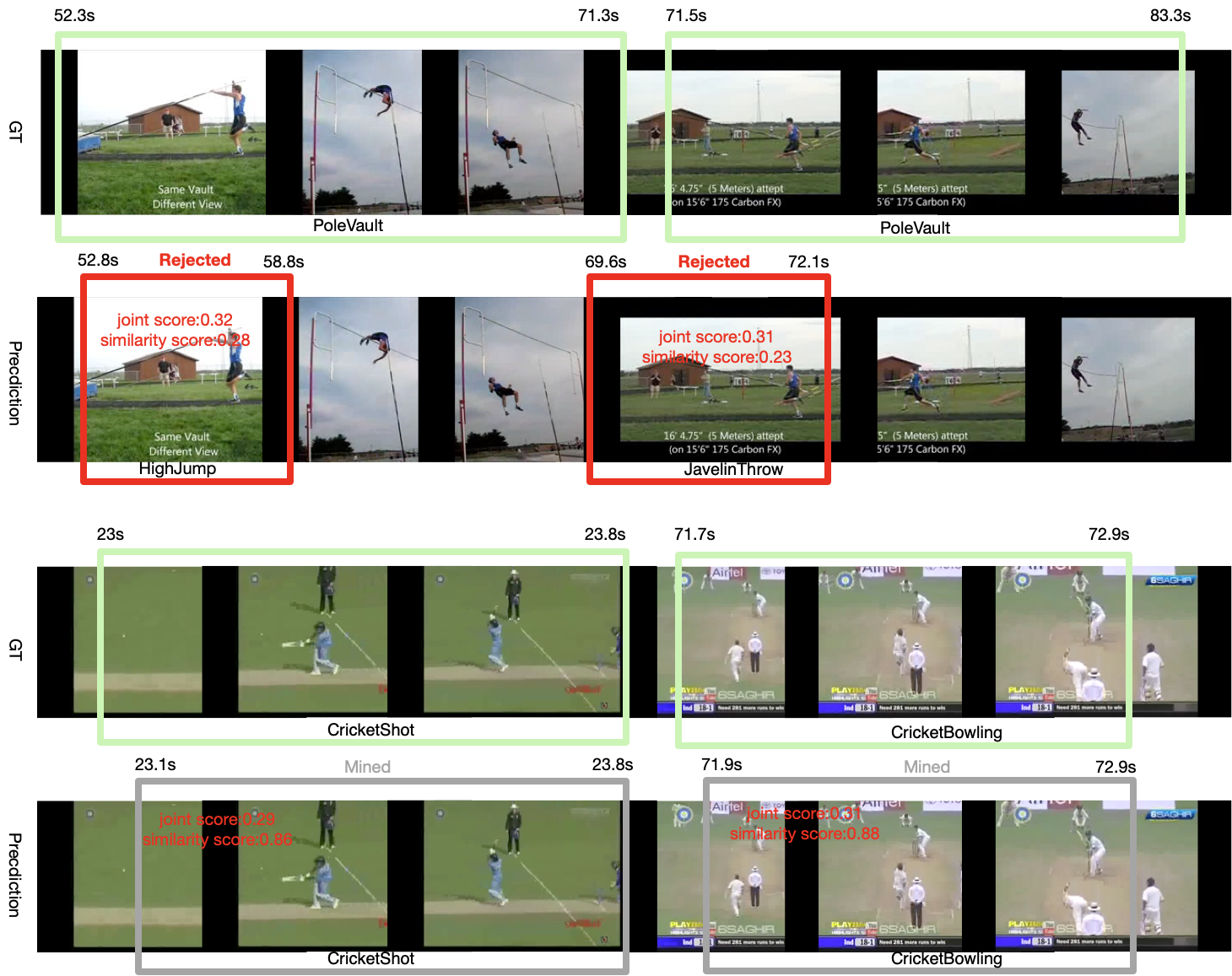}
\end{center}
\caption{Examples of rejected and mined positives on THUMOS14.}
\label{fig:example}
\end{figure*}

\textbf{Qualitative results.} To better illustrate the effectiveness of our method, we visualize some qualitative localization results on the THUMOS14 and ActivityNet v1.3 datasets in Fig. \ref{fig:vis_results}. Since the recent work NPL \cite{xia2023learning} does not provide codes and implementation details, we refrain from direct qualitative comparisons.  In addition, we also show some examples of rejected and mined positives on THUMOS14 during training, as shown in Fig. \ref{fig:example}. 

\end{document}